\documentclass[lettersize,journal]{IEEEtran}
\usepackage{amsmath,amsfonts}
\usepackage{algorithmic}
\usepackage{algorithm}
\usepackage{array}
\usepackage[caption=false,font=normalsize,labelfont=sf,textfont=sf]{subfig}
\usepackage{textcomp}
\usepackage{tabularx}
\usepackage{stfloats}
\usepackage{booktabs}
\usepackage{url}
\usepackage{verbatim}
\usepackage{graphicx}
\usepackage{cite}
\usepackage{pifont}
\newcommand{\cmark}{\ding{51}}
\usepackage{caption}
\usepackage[table]{xcolor}
\usepackage{makecell}
\usepackage{placeins}
\begin{document}

\title{RealEmbedBench: Benchmarking LLMs on Real-World Application-Layer Embedded Software Tasks}

\author{
\begin{center}
\begin{minipage}[t]{0.33\linewidth}
\centering
Yujun Huang\\
Nanjing University of Aeronautics and Astronautics\\
Nanjing, China\\
yujunhuang@nuaa.edu.cn
\end{minipage}%
\hfill
\begin{minipage}[t]{0.33\linewidth}
\centering
Yuwei Zhang\\
Nanjing University of Aeronautics and Astronautics\\
Nanjing, China\\
zyw.19850975279@gmail.com
\end{minipage}%
\hfill
\begin{minipage}[t]{0.33\linewidth}
\centering
Yi Zhou\\
Nanjing University of Aeronautics and Astronautics\\
Nanjing, China\\
\end{minipage}

\vspace{1em}

\begin{minipage}[t]{0.33\linewidth}
\centering
Shanshan Wu\\
Nanjing Research institute of electronics engineering\\
Nanjing, China\\
\end{minipage}%
\hfill
\begin{minipage}[t]{0.33\linewidth}
\centering
Xiang Chen\\
Nanjing University of Aeronautics and Astronautics\\
Nanjing, China\\
xiang\_chen@nuaa.edu.cn
\end{minipage}%
\hfill
\begin{minipage}[t]{0.33\linewidth}
\centering
Lina Gong*\\
Nanjing University of Aeronautics and Astronautics\\
Nanjing, China\\
gonglina@nuaa.edu.cn
\end{minipage}
\end{center}
}

\markboth{Journal of \LaTeX\ Class Files,~Vol.~14, No.~8, August~2021}%
{Shell \MakeLowercase{\textit{et al.}}: A Sample Article Using IEEEtran.cls for IEEE Journals}

\IEEEoverridecommandlockouts
\IEEEpubid{0000--0000/00\$00.00~\copyright~2021 IEEE}

\maketitle

\begin{abstract}
Large language models (LLMs) have achieved strong performance on general coding tasks, however, their capabilities in embedded software development remain underexplored due to the lack of dedicated benchmarks. Existing benchmarks for embedded development, such as EmbedAgent, primarily focus on hardware-in-the-loop validation in simulated environments, with limited attention to application-layer software that operates above the hardware abstraction layer. To address this gap, we propose RealEmbedBench, a benchmark covering three representative application-layer tasks: code generation and completion, state machine modeling and implementation, and cross-OS code translation, comprising a total of 5,142 samples. We further evaluate six representative LLMs on these tasks through extensive experiments. The results reveal substantial variation in performance across tasks and models. Project context has a significant impact on code generation and completion, particularly with respect to the code window. In the state machine tasks, models perform reasonably well at modeling state machines, but exhibit clear deficiencies when translating these models into executable code. Cross-OS translation tasks show pronounced directional differences, with model performance varying significantly across different translation directions, and certain directions remain particularly challenging for almost all models, such as from Harmony to RT-Thread, which has an average pass rate of only 0.1667. These findings suggest that developers should pay particular attention to the project context, especially the local code window, when using LLMs for code generation and completion, and select models according to the characteristics of each specific translation direction for cross-OS translation tasks. LLM trainers, meanwhile, should invest more in real-world state machine implementation data to bridge the gap between structural modeling and executable code implementation.
\end{abstract}

\begin{IEEEkeywords}
Embedded Systems,~Large Language Models,~Benchmark,~State Machine Modeling,~Cross-OS Translation,~Context Dependency
\end{IEEEkeywords}

\section{Introduction}
\label{intro}
\IEEEPARstart{L}{arge} language models have shown significant potential in software development. Embedded software, as a critical part of software development, plays an essential role in domains such as automotive systems, industrial control, and the Internet of Things. To evaluate the capabilities of large language models across different software development tasks, numerous benchmarks, including HumanEval~\cite{humaneval} and SWE-Bench~\cite{swebench}, have been proposed. However, these benchmarks primarily focus on general-purpose software and do not fully capture the unique requirements of embedded development, such as strict resource constraints, hardware interactions, and complex state logic. 

\IEEEpubidadjcol Recently, some benchmarks for LLMs in embedded systems have been proposed, but they mainly target hardware-related tasks or low-level software development. For example, EmbedAgent~\cite{EmbedAgent} evaluates LLMs with hardware-in-the-loop tasks on the Wokwi \cite{wokwi} platform. Englhardt et al.~\cite{englhardt:chi-2024} study their performance in low-level driver development with physical validation. These benchmarks pay little attention to application-layer software development, where developers often face more complex business logic, behavior design, and operating system ecosystems than in low-level driver or hardware-interaction tasks. 

\begin{figure*}[!t]
\centering
\includegraphics[width=0.95\textwidth]{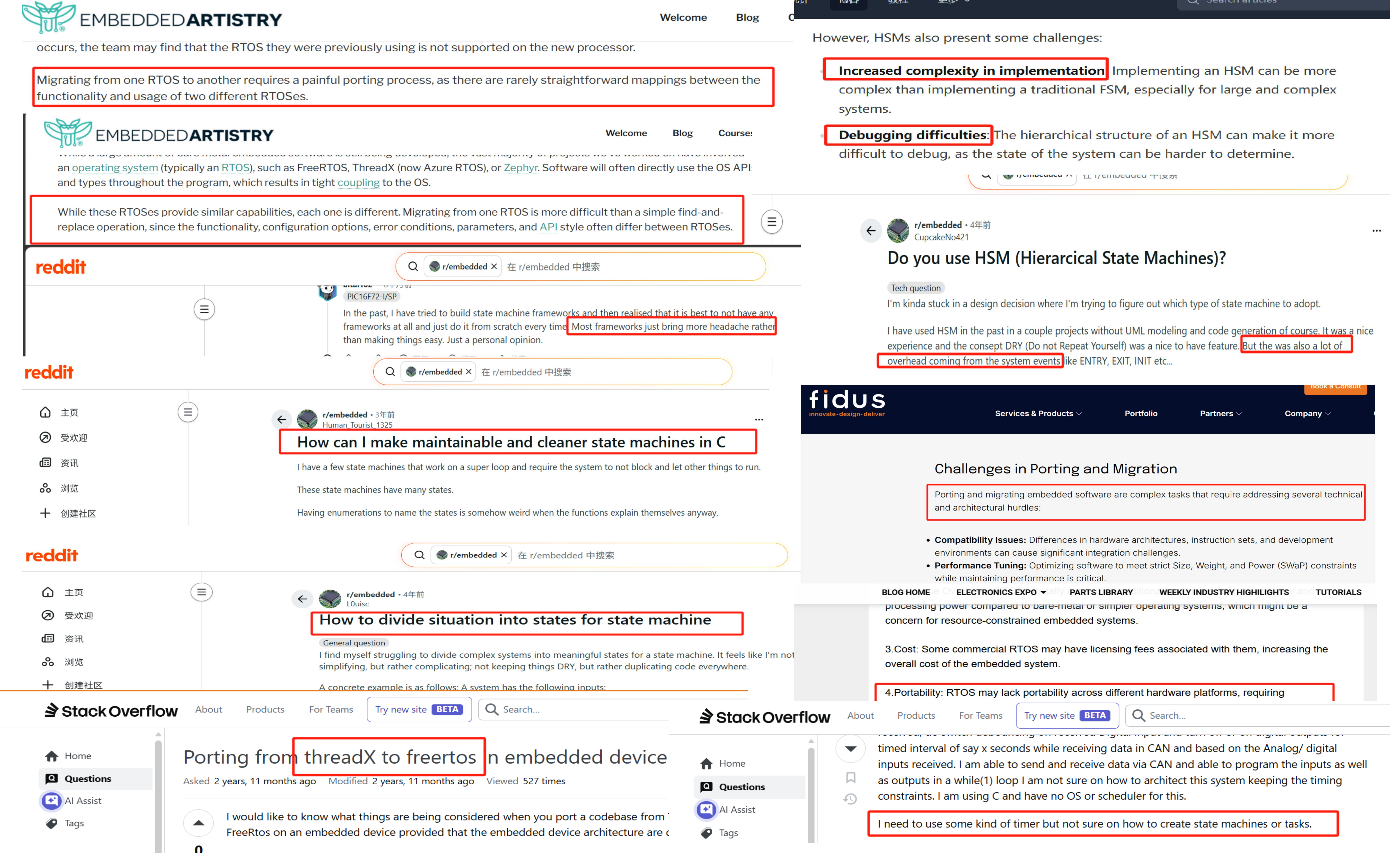}
\caption{Recurring pain points identified from embedded developer community discussions.}
\label{fig:user_study}
\end{figure*}

Inspired by the above observations, we first conducted a preliminary study by extensively collecting posts related to embedded application-layer development from platforms such as StackOverflow and Reddit. After filtering, 125 posts were retained (Figure~\ref{fig:user_study}); these posts were then labeled and categorized to capture the pain points and challenges developers face in embedded application-layer software development. We subsequently conducted semi-structured interviews with 12 developers to validate the findings from the online analysis. Building on these two stages, we identified and formally defined three primary categories of tasks: \textbf{code generation and completion in complex project contexts}, \textbf{state machine modeling and implementation}, and \textbf{cross-OS code translation}, which together cover the core challenges encountered in embedded application-layer software development.

Based on these task definitions, we propose a multi-stage benchmark construction framework to address this gap, consisting of \textbf{fundamental data collection}, \textbf{task-specific dataset construction}, and \textbf{benchmark statistics}. In the first stage, representative application domains are determined and diverse, high-quality projects are collected. The task-specific dataset construction stage then covers three primary tasks: \textbf{code generation and completion}, \textbf{state machine modeling and implementation}, and \textbf{cross-OS code translation}, with automatically constructed samples manually reviewed to ensure task-specific correctness. Using this framework, we build \textbf{RealEmbedBench}, which contains \textbf{5,142} samples in total, including \textbf{1,426} code generation samples, \textbf{3,026} code completion samples, \textbf{80} state machine samples, and \textbf{610} cross-OS code translation pairs, providing a comprehensive resource for evaluating LLM performance in embedded application-layer software development. Detailed construction procedures are described in Section~\ref{sec:benchmark_construction}.

Using \textbf{RealEmbedBench}, we conducted extensive evaluations on six representative large language models across the three task categories covered by the benchmark. We first compared the overall performance of all evaluated models on each task category and observed substantial differences across both models and tasks. Overall, \textbf{Gemini-Pro 3.1} achieved the best average performance. We then further explored the key factors influencing model performance in each task through ablation experiments and case analyses. For code generation and completion, the results show that project context plays a critical role in model performance, with nearby code context providing the most consistent benefit. For state machine tasks, the analysis indicates that the tasks remain challenging when models need to transform structural designs into faithful implementations. For cross-OS translation tasks, performance depends strongly on the specific source and target operating systems, with many failures caused by incomplete preservation of target-side semantics such as API behavior, parameter usage, and return conditions.

These findings provide further practical implications for the use and improvement of large language models in embedded software engineering. In code generation and completion tasks, project context is critical for model performance, particularly the nearby code window and call-chain information, which provide direct constraints on implementation and help models generate code consistent with file-level logic and execution flow. In cross-operating-system translation tasks, developers should pay attention to both source and target OS types, as well as common error patterns such as mismatched semantic granularity, incorrect parameter interpretation, and omission of implicit behavioral constraints, which can lead to code that appears reasonable superficially but behaves incorrectly under the target OS. For state-machine-related tasks, there remains a clear gap between structural understanding and faithful code implementation; model training should place greater emphasis on data that connects design semantics with concrete implementation details, including state-machine skeletons, condition-check placement, transition expansion patterns, and project-specific API conventions, to reduce implementation errors and better align generated code with embedded software practice.

The main contributions of this paper are as follows:
\begin{itemize}

\item We introduce \textbf{RealEmbedBench}, a benchmark for evaluating large language models in embedded application software development, together with a benchmark construction and evaluation framework. It contains \textbf{5,142} manually reviewed samples covering three representative task categories: code generation and completion, state machine modeling and implementation, and cross-OS code translation.The dataset and related code are publicly available on GitHub: \url{https://github.com/huangyujun1215/RealEmbedBench}.

\item We conduct systematic evaluations of six representative large language models on \textbf{RealEmbedBench}, together with ablation studies and error analysis. Our results reveal the critical role of project context in code generation, the gap between structural understanding and faithful implementation in state machine tasks, and the strong direction dependence of cross-OS translation performance.

\end{itemize}

\section{PRELIMINARY STUDY AND RELATED WORK}
\begin{figure*}[!t]
\centering
\includegraphics[width=1\textwidth]{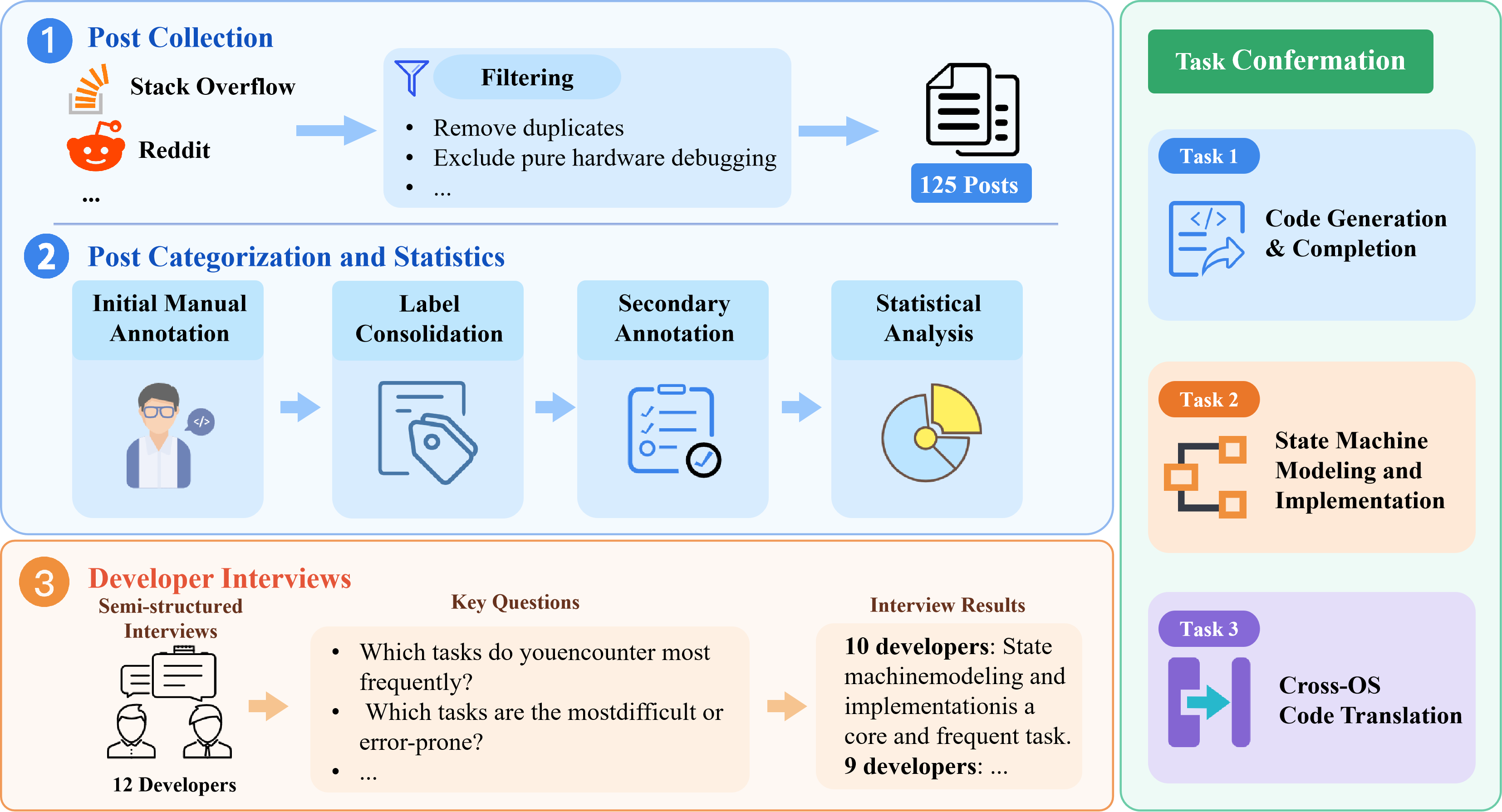}
\caption{Preliminary study pipeline}
\label{fig:pre_study}
\end{figure*}

In this section, we conduct a preliminary study based on community discussions and developer interviews to identify recurring challenges in application-layer embedded software development (Figure~\ref{fig:pre_study}). Based on the findings of this study, we then define the core tasks of RealEmbedBench and compare our benchmark with existing code-related benchmarks.

\subsection{Task Identification}

To ensure that the benchmark reflects practical embedded development needs rather than arbitrary task selection, we first investigate what challenges are repeatedly encountered by developers in real projects. We combine large-scale online discussion analysis with developer interviews, and use the resulting evidence to identify the most representative task categories for RealEmbedBench.

\subsubsection{Post collection.} 
\label{post_collection}
As shown in Part 1 of Fig.~\ref{fig:pre_study}, we collected discussions related to embedded software development from multiple online platforms, including Stack Overflow and Reddit, to obtain a broad view of practical developer needs. We focused on topics that reflect common developer challenges, such as optimizing firmware performance, managing RTOS tasks, implementing robust driver solutions, designing complex control algorithms, ensuring system integration, and enhancing software security. After filtering out duplicated posts, discussions unrelated to software development, and posts concerned only with pure hardware debugging, we retained \textbf{\textit{125}} posts for further analysis.

\subsubsection{Post categorization and statistics.}  
To systematically analyze the 125 retained posts, we adopted a multi-stage labeling and statistical analysis procedure, as illustrated in Part 2 of Fig.~\ref{fig:pre_study}

\begin{enumerate}
    \item \textbf{Initial Manual Annotation.} 
    Three graduate researchers independently conducted a cross-analysis of the collected posts. Each post was assigned a single label representing its primary software engineering issue, based on its keywords and discussion content.

    \item \textbf{Label Consolidation.} 
    Following the initial annotation, the preliminary labels obtained from the first round of annotation were reviewed and consolidated. Overlapping or semantically similar labels were merged, and a unified labeling taxonomy was established to ensure conceptual clarity and consistency in the classification scheme.

    \item \textbf{Secondary Annotation.} 
    Once the labels were refined, we proceeded with a secondary annotation. Using the refined taxonomy, all posts were re-annotated to ensure alignment with the standardized classification criteria. This second-round annotation improved labeling consistency and enhanced the reliability of the results.

    \item \textbf{Statistical Analysis.} 
    Finally, based on the finalized labels, we performed a frequency analysis of each category. The distribution and proportion of the identified categories are summarized in Table~\ref{tab:post_statistics}.
\end{enumerate}

\begin{table}[t]
\centering
\caption{Distribution of Embedded Software Engineering Issues in the Collected Posts}
\label{tab:post_statistics}
\footnotesize
\setlength{\tabcolsep}{3pt}
\renewcommand{\arraystretch}{1.08}
\begin{tabularx}{\columnwidth}{>{\raggedright\arraybackslash}Xcc}
\toprule
\textbf{Category} & \textbf{Count} & \textbf{Percentage} \\
\midrule
Context-aware code generation and completion & 33 & 26.4\% \\
State machine modeling and implementation & 28 & 22.4\% \\
Cross-OS code translation & 23 & 18.4\% \\
RTOS scheduling and task management & 15 & 12.0\% \\
Firmware--hardware coupling and abstraction design & 12 & 9.6\% \\
Embedded system architecture and modularization & 9 & 7.2\% \\
Others & 5 & 4.0\% \\
\midrule
\textbf{Total} & \textbf{125} & \textbf{100\%} \\
\bottomrule
\end{tabularx}
\end{table}

Based on the statistical results, we found that 
\textbf{context-aware code generation and completion}, 
\textbf{state machine modeling and implementation}, and 
\textbf{cross-OS code translation} were the three most frequent categories, accounting for 
\textbf{\textit{26.4\%}}, 
\textbf{\textit{22.4\%}}, and 
\textbf{\textit{18.4\%}} of the retained posts, respectively.

\subsubsection{Developer interviews.}
Building upon the findings from the online analysis, we subsequently conducted semi-structured interviews to validate whether these frequently discussed issues also reflect industrial practice. The interview protocol was designed based on the categories identified in the post analysis phase(see Part 3 of Fig.~\ref{fig:pre_study}). In particular, we asked participants to describe the tasks they encounter most frequently, the tasks they consider difficult or error-prone, and whether activities such as state machine modeling and implementation, context-aware code generation \& completion, and RTOS translation are common in their daily work.

The interview results are largely consistent with the trends observed in the online analysis. Among the 12 participants, 10 explicitly reported that \textbf{state machine modeling and implementation} is a core and frequently encountered task in their daily development, particularly in safety-critical and event-driven systems. Similarly, 9 participants mentioned \textbf{context-aware code generation and completion} as a highly desirable capability, especially when working with large legacy codebases and complex project structures. In addition, 8 developers indicated that \textbf{cross-OS code translation} or RTOS translation is a recurring and non-trivial activity in their projects.

\subsubsection{Task Confirmation.}
Based on the above evidence, we formally establish these three task categories as the foundation of our benchmark. 

\noindent\textbf{Task 1: code generation and completion.}

\textbf{Description.}
This task generally refers to two aspects: generating executable code fragments or complete implementations based on natural language descriptions and completing partial code contexts by filling in missing parts to ensure continuity and consistency within the existing code structure. Such code generation or completion tasks have been widely studied in general software  development~\cite{humaneval,austin2021program,lu2021codexglue}. However, in embedded software development, the task exhibits stronger system-level dependencies and engineering constraints. In practice, embedded code generation commonly involves device drivers, task control logic, initialization procedures, interrupt handlers, and hardware- or RTOS-related interfaces~\cite{schirner2007embedded,garousi2018testing}. Compared with general-purpose software, embedded systems are constrained by limited resources, real-time requirements, and platform-specific interfaces. Consequently, generated code must satisfy not only functional correctness but also interface compatibility, timing constraints, and engineering usability.~\cite{eclipse-iot-survey-2024}.

\medskip
\textbf{Definition.}
In this work, we formulate embedded software development as two related tasks: code generation and code completion. Let \(C\) denote the carefully curated and categorized development context, \(S\) the task specification, \(P\) the partial code context, and \(L\) the large language model.

For code generation, the target code \(Y_{\text{gen}}\) is defined as
\begin{equation}
Y_{\text{gen}} = L(C, S).
\label{eq:code_generation}
\end{equation}

For code completion, the target code \(Y_{\text{comp}}\) is defined as
\begin{equation}
Y_{\text{comp}} = L(C, S, P).
\label{eq:code_completion}
\end{equation}

To systematically characterize the context provided to the model, we first adopt a classical syntactic classification of code context from software engineering literature~\cite{schirner2007embedded,yang2024embedgenius}. Based on this structural categorization, we extract key modules and source files from the embedded software projects listed in Table~\ref{tab:embedded_projects} and perform sample analysis. During this process, we systematically identify the concrete engineering constructs corresponding to each syntactic context type. For example, local code windows capture state checks, buffer operations, and sequential control logic; import dependencies correspond to HAL headers, protocol libraries, and driver interfaces; call dependencies include interrupt handlers, message dispatch routines, and RTOS task callbacks; global data represents device flags, control blocks, and inter-task communication buffers; and type definitions cover message structs, state enums, and configuration structs. Table~\ref{tab:context_pattern_mapping} summarizes the mapping between context types and their typical content patterns in embedded software projects. This structured context modeling provides a clear and operational foundation for task formulation.
\medskip
\medskip

\begin{table}[!t]
\centering
\caption{Mapping between syntactic context types and representative embedded engineering content}
\label{tab:context_pattern_mapping}
\scriptsize
\setlength{\tabcolsep}{3pt}
\renewcommand{\arraystretch}{1.08}
\begin{tabularx}{\columnwidth}{
>{\raggedright\arraybackslash}p{0.34\columnwidth}
>{\raggedright\arraybackslash}X}
\toprule
\textbf{Syntactic Context Type} & \textbf{Mapped Embedded Software Content} \\
\midrule
Local code window & State checks, buffer operations, sequential logic \\
\midrule
Import dependencies & HAL headers, protocol libraries, driver interfaces \\
\midrule
Call dependencies & Interrupt handlers, message dispatch, task callbacks \\
\midrule
Global data & Device flags, control blocks, global buffers \\
\midrule
Type definition & Message structs, state enums, configuration structs \\
\bottomrule
\end{tabularx}
\end{table}

\noindent\textbf{Task 2: State Machine Modeling and Implementation.}

\textbf{Description.}
State machines constitute a fundamental paradigm in reactive embedded software, where system behavior is structured as discrete states and event-triggered transitions~\cite{yang2021single,yang2020painting}. This modeling approach is particularly critical in embedded systems that must respond deterministically to asynchronous inputs, hardware interrupts, and timing events under strict resource and real-time constraints~\cite{yang2024embedgenius,schirner2007embedded}. State machines are also central to model-driven development workflows, in which system requirements are progressively refined into behavioral models and subsequently transformed into executable implementations~\cite{wkasowski2005code}. As a result, embedded state machine development often involves multiple interconnected representations, including natural language specifications, PlantUML-based state machine models, and source code implementations. However, current benchmarks rarely test LLMs on embedded state machine development across these representations.

\medskip
\textbf{Definition.}
In this work, we formulate embedded state machine development as a set of cross-representation transformation tasks. Let \(S_{\mathrm{NL}}\) denote a state machine specification in natural language, \(S_{\mathrm{PUML}}\) a state machine model expressed in PlantUML, \(S_{\mathrm{Code}}\) the corresponding source code implementation, \(C\) the embedded development context, and \(L\) the large language model.

The natural-language-to-PlantUML transformation is defined as
\begin{equation}
S_{\mathrm{PUML}} = L(C, S_{\mathrm{NL}}).
\label{eq:nl_to_puml}
\end{equation}
This task evaluates whether the model can accurately interpret natural language requirements and derive a state machine model in PlantUML.

The PlantUML-to-code transformation is defined as
\begin{equation}
S_{\mathrm{Code}} = L(C, S_{\mathrm{PUML}}).
\label{eq:puml_to_code}
\end{equation}
This task evaluates whether the model can translate a PlantUML state machine model into an executable implementation in embedded software.

The natural-language-to-code transformation is defined as
\begin{equation}
S_{\mathrm{Code}} = L(C, S_{\mathrm{NL}}).
\label{eq:nl_to_code}
\end{equation}
This task evaluates whether the model can directly synthesize executable state machine code from natural language specifications.

Together, these three formulations assess the ability of LLMs to consistently interpret, transform, and implement embedded state machine logic across heterogeneous representations while preserving behavioral semantics.

\medskip
\medskip

\noindent\textbf{Task 3: Cross-OS Translation.}

\textbf{Description.}
This task focuses on translating software implemented for one operating system into a functionally equivalent implementation for another. In embedded software development, such scenarios commonly arise when application logic must be migrated across real-time operating systems (RTOSes) such as FreeRTOS, Zephyr, RT-Thread, and HarmonyOS. Although the high-level functionality may remain unchanged, the translation process typically requires adapting task management mechanisms, synchronization primitives, timing services, memory handling strategies, and other OS-specific interfaces~\cite{abdelsamea2016real,collie2020m3}.Because embedded implementations are tightly coupled with low-level system abstractions, differences in APIs, scheduling semantics, resource management policies, and error handling mechanisms can significantly affect translation correctness. Prior work has explored portability support and systematic code adaptation for embedded platforms~\cite{antonio2012systematic,xin2024firmporter}. However, current LLM benchmarks rarely evaluate whether models can perform such cross-OS adaptation under realistic embedded constraints.

\medskip
\textbf{Definition.}
In this work, we formulate embedded operating system translation as a cross-platform code translation task. Let \(S_{\mathrm{src}}\) denote an implementation targeting the source operating system \(O_{\mathrm{src}}\), \(S_{\mathrm{tgt}}\) the translated implementation targeting the target operating system \(O_{\mathrm{tgt}}\), \(C\) the embedded development context, and \(L\) the large language model.

The task is defined as
\begin{equation}
S_{\mathrm{tgt}} = L(C, S_{\mathrm{src}}, O_{\mathrm{src}}, O_{\mathrm{tgt}})
\label{eq:os_translation}
\end{equation}
subject to
\begin{equation}
\mathrm{Sem}(S_{\mathrm{tgt}}, O_{\mathrm{tgt}}) \equiv \mathrm{Sem}(S_{\mathrm{src}}, O_{\mathrm{src}}),
\label{eq:os_equivalence}
\end{equation}
where \(\mathrm{Sem}(\cdot)\) denotes the program behavior under the corresponding operating system.

Equation~\ref{eq:os_translation} represents the generation of a target implementation for the target operating system, while Equation~\ref{eq:os_equivalence} requires the translated implementation to preserve the functional behavior of the original program.

\subsection{Benchmarks for Evaluating LLMs on Code-Related Tasks}
Having defined the three core tasks targeted by RealEmbedBench, we next position our work within the landscape of existing LLM evaluation benchmarks.
Recent work has introduced many benchmarks for evaluating LLMs on code-related tasks, covering function-level code generation, competitive programming, data science code generation, and more realistic software engineering tasks such as debugging, repair, and editing~\cite{humaneval,austin2021program,hendrycks2021measuring,lai2023ds,lu2021codexglue,swebench,tian2024debugbench,guo2024codeeditorbench,zeng2024coderujb}. However, most of these benchmarks focus on general programming scenarios and do not adequately reflect the characteristics of embedded software development, such as hardware constraints, real-time requirements, RTOS diversity, and state-machine-based control logic. To address this gap, RealEmbedBench focuses on application-layer embedded software development and evaluates three underexplored capabilities: context-aware coding in realistic embedded projects, state machine modeling and implementation across multiple representations, and cross-OS code translation under heterogeneous system abstractions. Table~\ref{tab:benchmark-comparison} compares the coverage of existing benchmarks with that of RealEmbedBench.

\begin{table}[!t]
\centering
\caption{Comparison of Existing Benchmarks for General Coding / Embedded Software Development}
\label{tab:benchmark-comparison}
\resizebox{\columnwidth}{!}{%
\begin{tabular}{@{}lcccc@{}}
\toprule
\textbf{Benchmark} & \textbf{\begin{tabular}[c]{@{}c@{}}Embedded\\ Focus\end{tabular}} & \textbf{\begin{tabular}[c]{@{}c@{}}Context\\ \end{tabular}} & \textbf{\begin{tabular}[c]{@{}c@{}}Cross-OS\\ Translation\end{tabular}} & \textbf{\begin{tabular}[c]{@{}c@{}}State Machine\\ Comprehension\end{tabular}} \\
\midrule
Englhardt et al.~\cite{englhardt:chi-2024} & \cmark & $\times$ & $\times$ & $\times$ \\
EmbedAgent~\cite{EmbedAgent} & \cmark & \cmark & $\times$ & $\times$ \\
BigCodeBench~\cite{bigcodebench} & $\times$ & \cmark & $\times$ & $\times$ \\
CoCo-Bench \cite{cocobench} & $\times$ & \cmark & $\times$ & $\times$ \\
Deveval \cite{li2024deveval} & $\times$ & \cmark & $\times$ & $\times$ \\
RefactorBench \cite{gautam2025refactorbench} & $\times$ & \cmark & $\times$ & $\times$ \\
Debugbench \cite{tian2024debugbench} & $\times$ & \cmark & $\times$ & $\times$ \\
\midrule
\textbf{RealEmbedBench (Ours)} & \cmark & \cmark & \cmark & \cmark \\
\bottomrule
\end{tabular}%
}
\end{table}

\section{Benchmark Construction}
\label{sec:benchmark_construction}
\begin{figure*}[!t]
\centering
\includegraphics[width=1\textwidth]{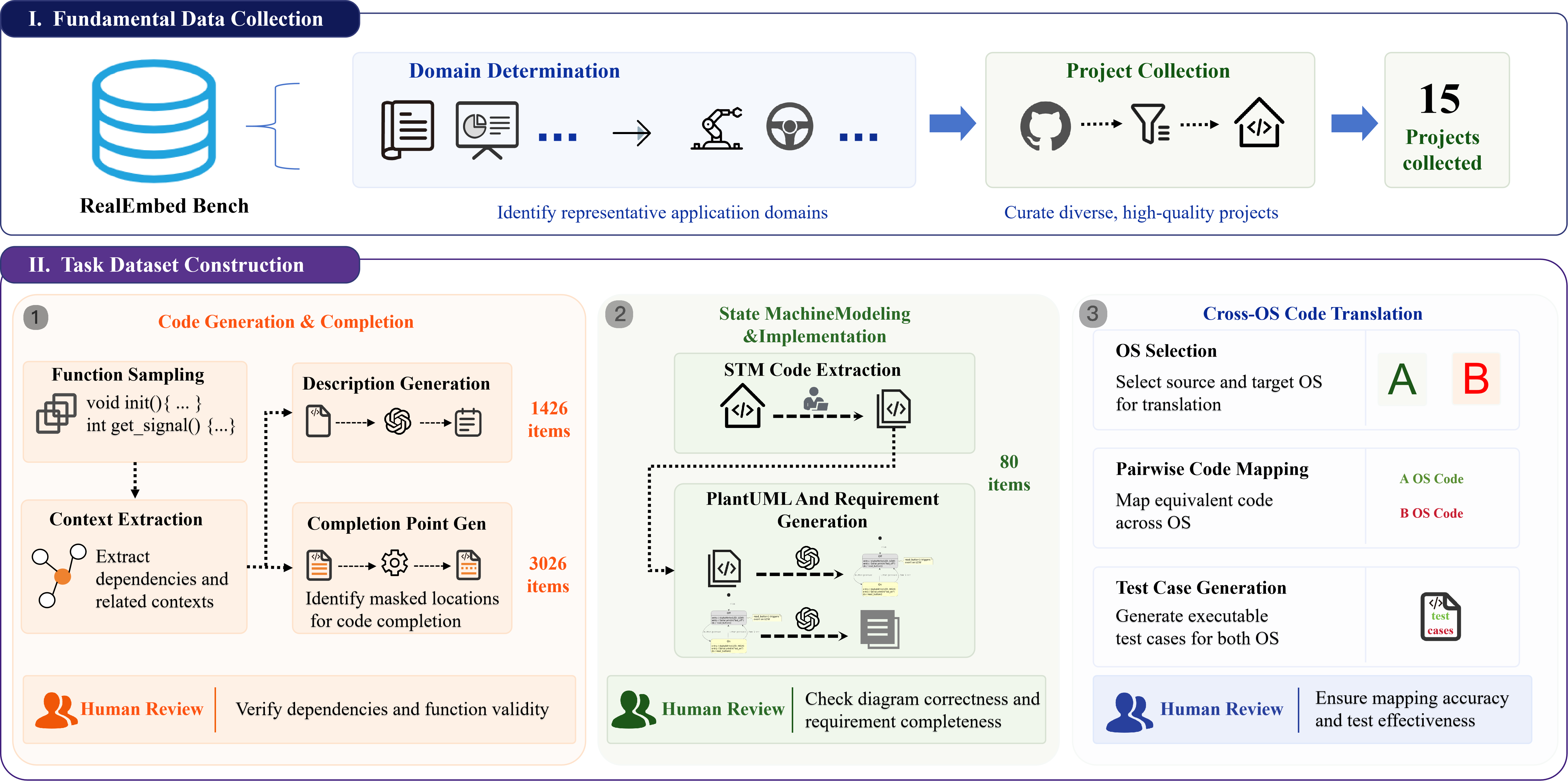}
\caption{RealEmbedBench data construction pipeline}
\label{fig:overview}
\end{figure*}

In this section, we describe the construction of RealEmbedBench, including fundamental data collection, task-specific dataset construction, and benchmark statistics. An overview of the pipeline is shown in Figure~\ref{fig:overview}.

\subsection{Fundamental Data Collection}
As shown in Part I of Figure~\ref{fig:overview}, this stage involves domain determination and project collection for benchmark construction, thereby establishing the foundation for the subsequent data processing, task formulation, and evaluation pipeline.

\textbf{Domain Determination.} To ensure the representativeness of our benchmark, we first considered the broad diversity of embedded software, which spans numerous application scenarios and industrial sectors. Since it is impractical to cover all possible subdomains, we identified representative embedded software subdomains by analyzing industry reports~\cite{eclipse-iot-survey-2024} and recent academic studies~\cite{garousi2018testing}. During this process, we excluded domains with limited open-source availability or heavy reliance on proprietary software, as such domains are less suitable for large-scale benchmark construction and reproducible evaluation. Following this selection procedure, we retained \textbf{\textit{10}} representative subdomains and collected corresponding open-source projects, as summarized in Table~\ref{tab:embedded_projects}.

\textbf{Project Collection.} Within these selected subdomains, we used an automated pipeline to mine candidate projects from GitHub and identify representative open-source projects for benchmark construction. For each subdomain, we first ranked candidate projects by popularity, using stars and forks as the primary criteria. We then filtered the ranked projects by maintenance status, retaining only those with recent update activity. This filtering step served two purposes: ensuring that the selected projects were still actively maintained and reducing the risk of data leakage from outdated or heavily memorized benchmark sources. Based on this process, we selected 15 large, actively maintained, and widely used open-source projects. As shown in Table~\ref{tab:embedded_projects}, these projects together form the source corpus for subsequent task construction.

\begin{table}[tbp]
\centering
\renewcommand{\arraystretch}{1.5} 
\caption{Embedded Systems Projects Overview}
\label{tab:embedded_projects}
\scriptsize
\newcolumntype{P}{>{\raggedright\arraybackslash\hsize=0.2\hsize}X}
\newcolumntype{D}{>{\raggedright\arraybackslash\hsize=0.25\hsize}X}
\newcolumntype{L}{>{\raggedright\arraybackslash\hsize=0.25\hsize}X}
\newcolumntype{S}{>{\raggedleft\arraybackslash\hsize=0.15\hsize}X}
\newcolumntype{F}{>{\raggedleft\arraybackslash\hsize=0.15\hsize}X}
\begin{tabularx}{\columnwidth}{@{} P D L S F @{}}
\toprule
\textbf{Project Name} & \textbf{Domain} & \textbf{Primary Language} & \textbf{Stars} & \textbf{Forks} \\
\midrule
ardupilot & Transportation & C/C++, Python, Objective-C & 14.2k & 19.9k \\
Marlin & Industrial automation & C/C++, Python & 17.2k & 19.6k \\
navigation2 & Industrial automation & C++, Python, CMake & 3.7k & 1.7k \\
openpilot & Automotive & Python, C++, Shell & 59.4k & 10.5k \\
RetroArch & Gaming & C, C++ & 12.4k & 2k \\
Tasmota & Home automation & C/C++, Berry, Python & 23.9k & 5k \\
tinyusb & Utilities & C & 6.3k & 1.3k \\
LVGL & General & C/C++, Python & 22.3k & 4k \\
littlefs & General & C, Python & 6.3k & 948 \\
esphome & Home automation & C/C++, Python & 10.2k & 4.9k \\
Thingsboard-Gateway & Internet of Things & Python & 2k & 955 \\
Wireless\-Android\-Auto\-Dongle & Automotive & C++ & 2.3k & 200 \\
openems & Energy management & Java/TypeScript & 1.2k & 534 \\
farmbot\_os & Agriculture & TypeScript, Ruby & 1.1k & 278 \\
open62541 & Industrial automation & C & 3k & 1.4k \\
\bottomrule
\end{tabularx}
\end{table}

\subsection{Code Completion and Generation}
As illustrated in Part II-1 of Figure~\ref{fig:overview}, this stage constructs the dataset for code completion and generation through function sampling, context extraction, description generation, and completion point generation.

\textbf{Function Sampling.} Our code generation and code completion tasks are constructed based on functions, so we first build a representative pool of functions from the collected projects before deriving benchmark instances. To reduce bias caused by variations in function length, we apply function-level stratified sampling. Specifically, we partition functions into five LOC-based intervals, $[0,20]$, $(20,50]$, $(50,100]$, $(100,200]$, and $[200,+\infty)$, with target sampling ratios of 20\%, 30\%, 30\%, 15\%, and 5\%, respectively. We adopt this partitioning and ratio design to better reflect the empirical distribution of functions in real-world embedded software, where short and medium-length functions are more prevalent, while long functions are less frequent but still important to retain for representativeness. The finer-grained intervals in the lower and middle LOC ranges also help preserve diversity among the most common function sizes. When a stratum contains insufficient candidates, the remaining quota is redistributed proportionally among the other strata. This process produces the foundational function set for subsequent code generation and code completion task construction.

\begin{figure}[!t]
\centering
\includegraphics[width=\columnwidth]{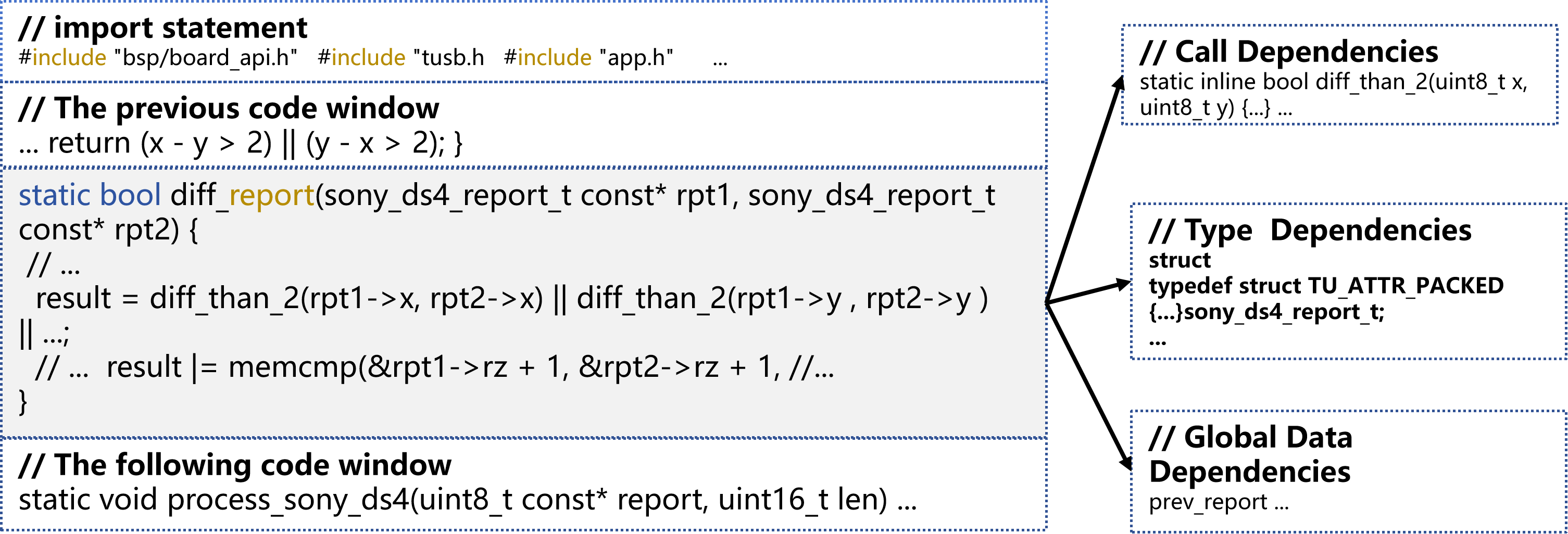}
\caption{Context classification}
\label{fig:dependency}
\end{figure}

\textbf{Context Extraction.} Based on the context categorization defined earlier in Table~\ref{tab:context_pattern_mapping}, we collected the corresponding contextual information required for each sampled function. For each function, we used Clang-based static analysis to extract its direct contextual dependencies, including called functions, accessed global variables, and referenced custom types such as structs, unions, enums, and typedefs. These extracted dependencies were then organized into the predefined context types shown in Figure~\ref{fig:dependency}. As illustrated in Figure~\ref{fig:dependency}, the context of an embedded function typically consists of multiple complementary parts: import statements provide external dependencies required for compilation, including board-support packages, hardware abstraction layers, protocol libraries, or application headers; the surrounding code windows capture local implementation flow before and after the target function, which helps reveal sequential logic and nearby usage patterns; call dependencies expose helper functions or utility routines invoked by the target function; type dependencies provide the data structures and type aliases needed to interpret device data, protocol messages, or memory layouts; and global data dependencies reveal shared states or buffers that the function reads or updates. Together, these context elements provide a more complete representation of the implementation environment in which an embedded function operates.

Since static analysis in embedded C/C++ can miss dependencies introduced by macros, conditional compilation, or project-specific build configurations, we further performed dedicated manual verification. Two reviewers manually checked all extracted dependencies, corrected more than 320 missing context items caused by static-analysis limitations, and removed more than 130 trivial boilerplate functions. This curation step improved the completeness and usefulness of the retrieved context for both the code generation and code completion tasks.

\begin{figure}[!t]
\centering
\includegraphics[width=\columnwidth]{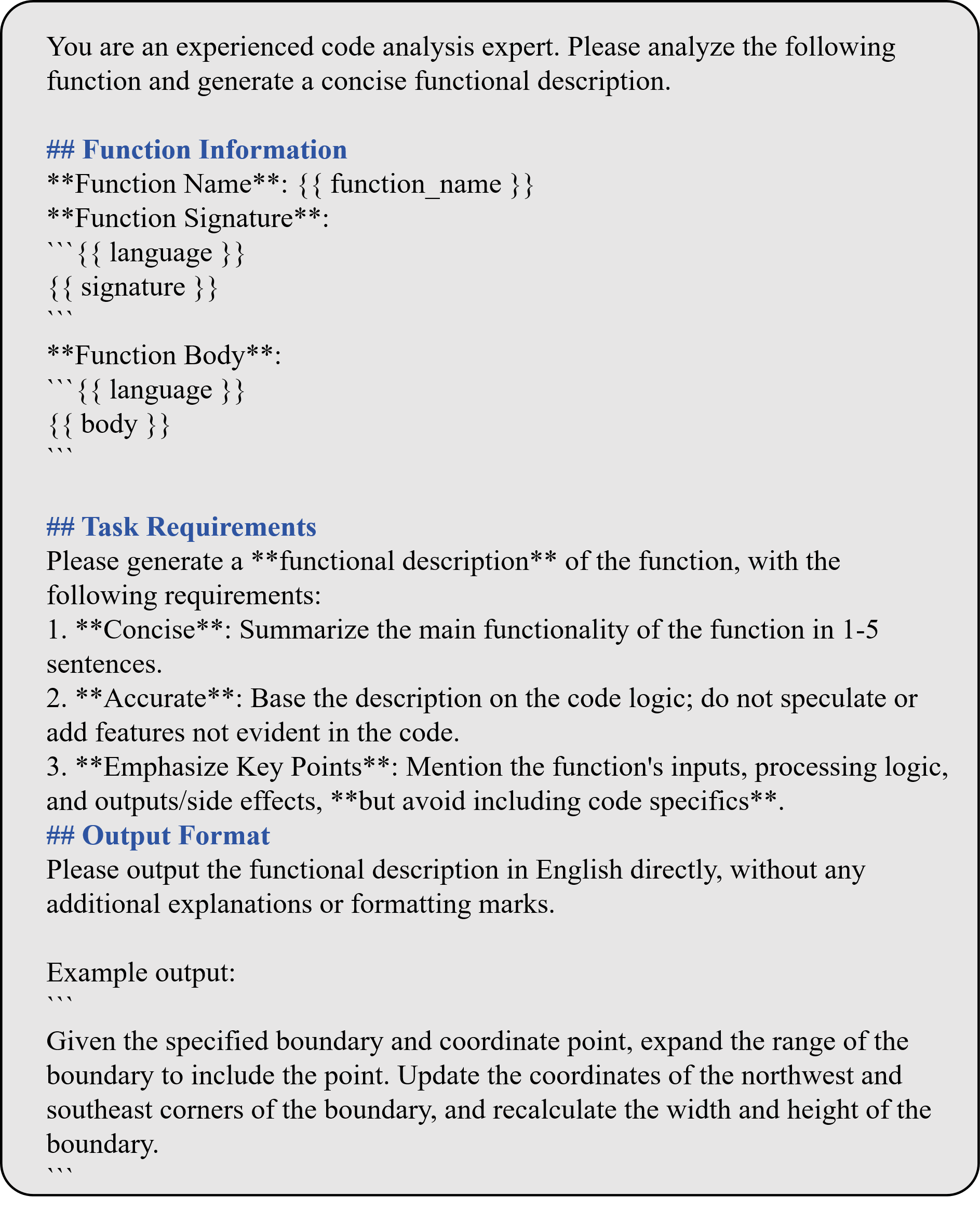}
\caption{Description Generation Prompt}
\label{fig:func_desc_prompt}
\end{figure}

\textbf{Description Generation.} For the code generation task, we follow the formulation in Equation~\ref{eq:code_generation}, where the model generates code $Y_{\text{gen}}$ from context $C$ and specification $S$. Under this formulation, the natural-language description serves as part of the specification $S$. We therefore used DeepSeek-v3.2 to generate a functional intent description for each sampled function based on the function code and its extracted context, resulting in 1,426 samples. The prompt template used for natural-language description generation is shown in Figure~\ref{fig:func_desc_prompt}. We then manually reviewed the generated descriptions to verify their correctness, completeness, and consistency with the original function semantics.

\textbf{Completion Point Generation.} For the code completion task, we follow the formulation in Equation~\ref{eq:code_completion}, where the LLM predicts the target code $Y_{\text{comp}}$ from context $C$, specification $S$, and partial code context $P$. To instantiate this formulation in our dataset, we construct $P$ and $Y_{\text{comp}}$ by masking a portion of the original function: the remaining visible code forms the partial code context $P$, while the masked segment is treated as the ground-truth target $Y_{\text{comp}}$ to be recovered by the model. To ensure that completion performance can be evaluated at different granularities, we apply masking at two levels: token-level masking for short-range completion and block-level masking for larger semantic units. Within each granularity setting, the masking position is selected randomly, reflecting the practical scenario in which code completion may be triggered at arbitrary points during development. This process yields 3,026 samples.

\subsubsection{State Machine Modeling and Implementation}
As illustrated in Part II-2 of Figure~\ref{fig:overview}, this phase constructs the state machine dataset through STM code extraction and PlantUML and requirement generation.

\textbf{STM Code Extraction.} Since state machine code is often tightly coupled with project-specific implementations, it is not straightforward to extract it independently and accurately from real-world projects. However, state machine implementations in practice usually follow several common patterns, such as \texttt{switch-case}-based state schedulers, lookup-table-based transition logic, and object-oriented state patterns~\cite{yang2021single,yang2020painting,wkasowski2005code,antonio2012systematic}. Based on this observation, we first applied heuristic static analysis to search for code fragments matching these patterns and treated them as candidate state-machine-related code. We then introduced manual inspection to examine these candidates and manually extracted the code confirmed to be related to state machine implementations.

In this process, we focused on state machine implementations with an explicit main loop, as this loop typically captures the core state transition logic. Accordingly, we divided each extracted state machine sample into two parts: the main loop code and its context, where the context consists of the single-layer dependencies required to implement the main loop. This design is also consistent with the task formulations in Equations~\ref{eq:puml_to_code} and~\ref{eq:nl_to_code}, where the target code \(S_{\mathrm{Code}}\) specifically corresponds to the extracted state machine main loop, while the required supporting dependencies are provided as part of the context \(C\). In the subsequent experiments, our goal is to evaluate the ability of LLMs to generate the main loop of a state machine. Therefore, we provide the dependencies necessary for generating the main loop in the prompt, so as to reduce the influence of project-specific dependencies on the results. Through this process, we finally obtained a collection of state machine code samples.

\begin{figure}[!t]
\centering
\includegraphics[width=\columnwidth]{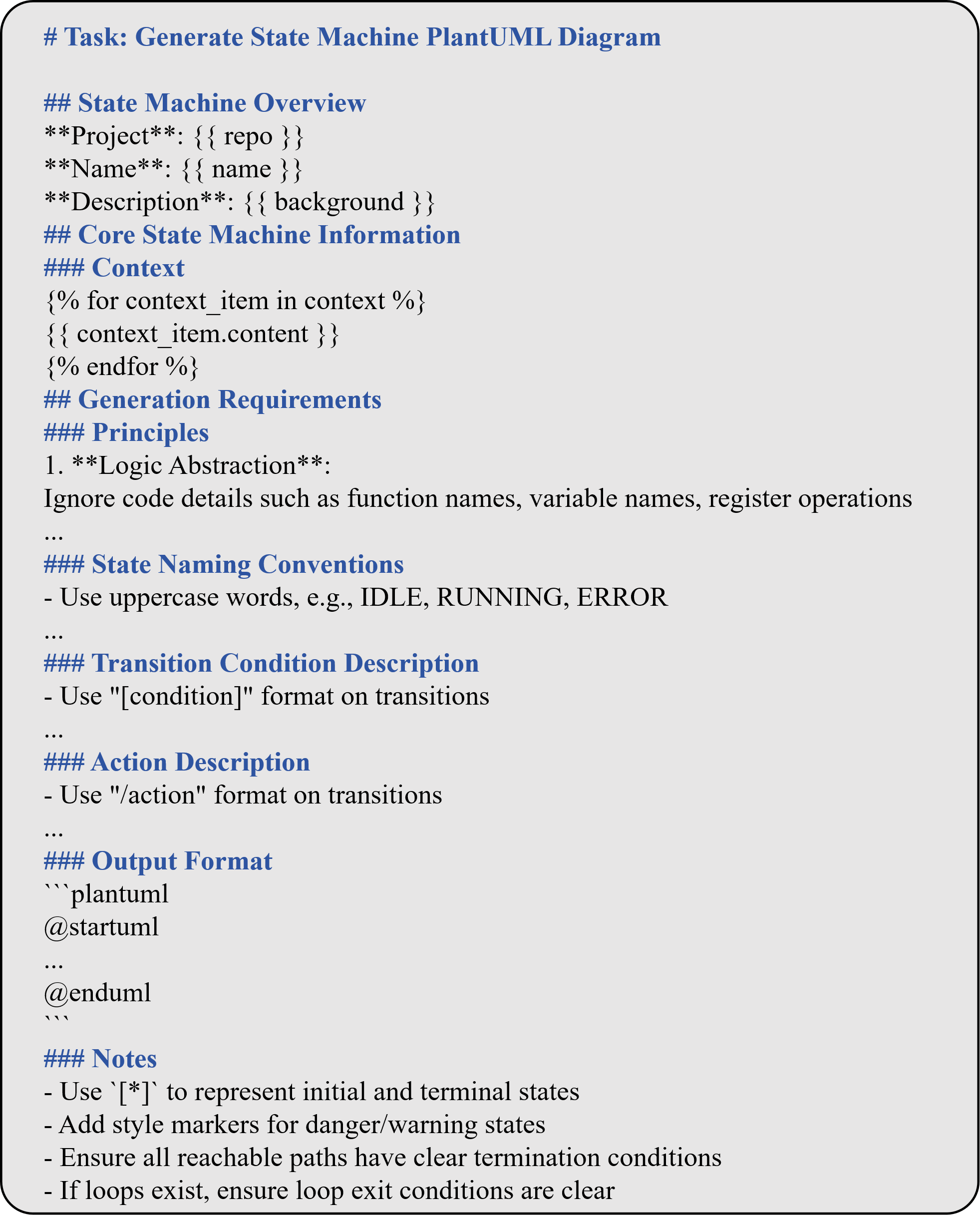}
\caption{PlantUML Generation Prompt}
\label{fig:generate_plantuml_prompt}
\end{figure}

\begin{figure}[!t]
\centering
\includegraphics[width=\columnwidth]{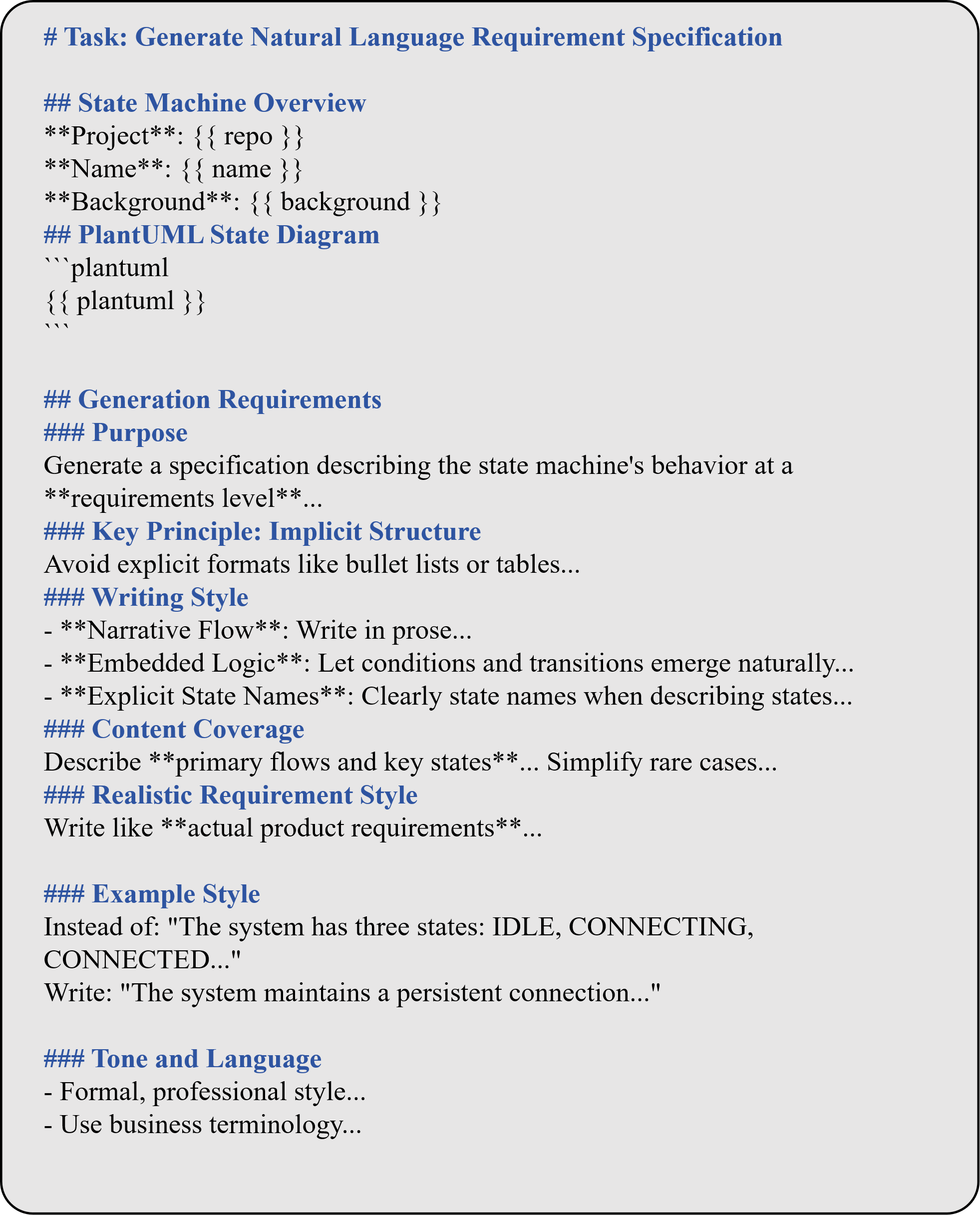}
\caption{Requirement Generation Prompt}
\label{fig:generate_requirement_prompt}
\end{figure}

\textbf{PlantUML And Requirement Generation.} In real-world embedded projects, the design and implementation of state machines typically follow a development process that starts from requirements, proceeds to formal modeling, and then moves to code implementation. To construct data that realistically reflects this workflow, we took the previously extracted collection of state machine code as the basis and further used DeepSeek-v3.2 to reversely derive the corresponding formal models and natural language requirements. Specifically, for the code representation \(S_{\mathrm{Code}}\) defined earlier, we first generated its corresponding PlantUML state machine model \(S_{\mathrm{PUML}}\), and then derived the associated natural language requirements \(S_{\mathrm{NL}}\) from the generated model. The prompts used for PlantUML generation and natural language requirement generation are shown in Figures~\ref{fig:generate_plantuml_prompt} and~\ref{fig:generate_requirement_prompt}, respectively.

\begin{figure}[!t]
\centering
\includegraphics[width=\columnwidth]{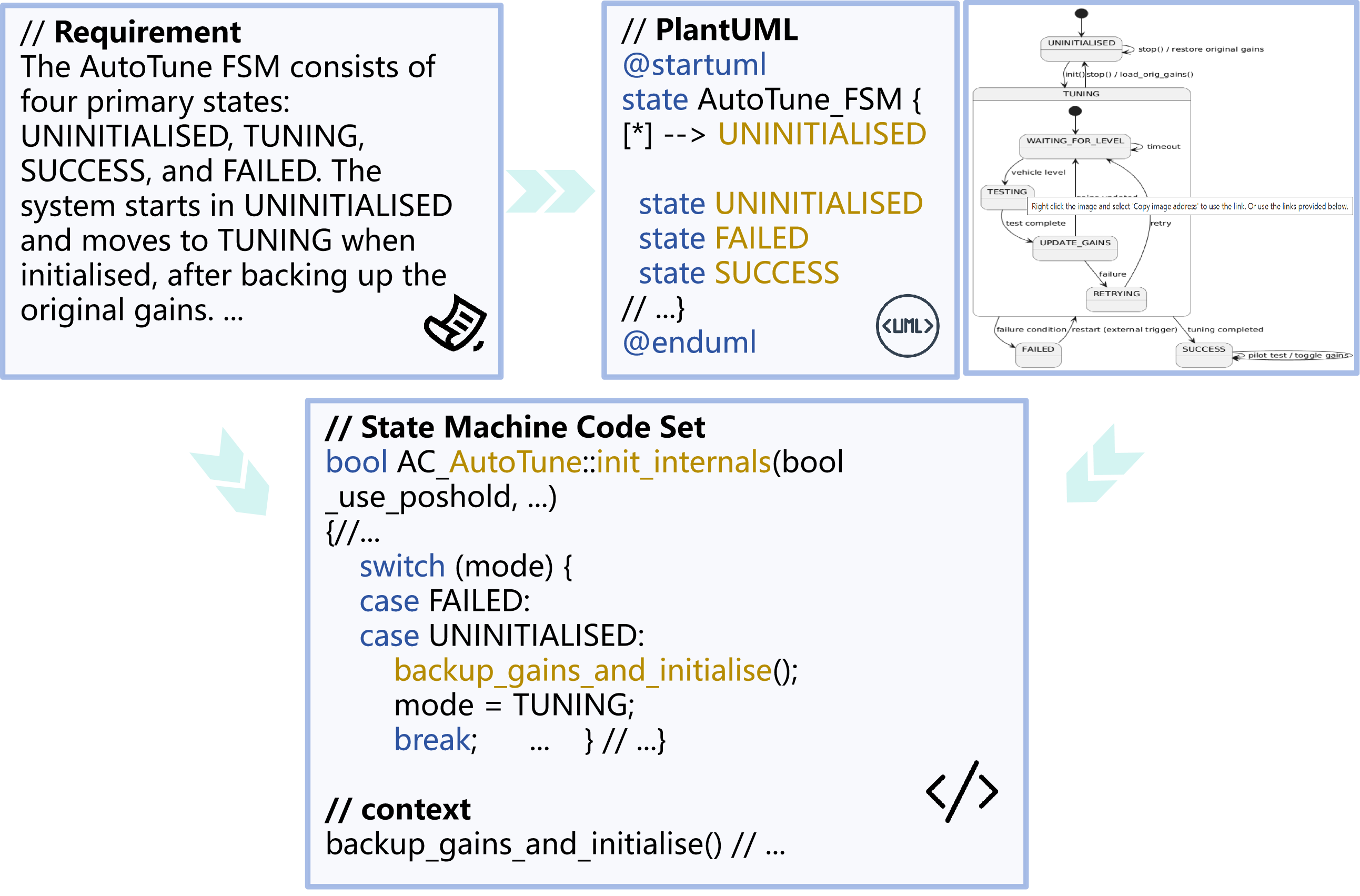}
\caption{State machine data example}
\label{fig:stm_example}
\end{figure}

Figure~\ref{fig:stm_example} presents a representative example of \texttt{AutoTune\_FSM}, illustrating the alignment among natural language requirements, the PlantUML state machine model, and the corresponding source code implementation. To ensure consistency across the three modalities, two reviewers manually examined the correspondence among the source code, generated PlantUML models, and natural language requirements, and corrected identified issues, including inconsistencies between models and requirements as well as missing boundary conditions. After revision, the consistency across the three modalities exceeded 98\%.

Based on these verified samples, we finally obtained 80 tri-modal samples consisting of natural language requirements, formal models, and source code, and constructed task datasets corresponding to Equations~\ref{eq:nl_to_puml}, \ref{eq:puml_to_code}, and \ref{eq:nl_to_code}.

\subsection{Cross-OS Code Translation}
As illustrated in Part II-3 of Figure~\ref{fig:overview}, this phase constructs the task data through abstract mining, pairwise code mapping, and test case generation.

\textbf{OS selection.} Based on the posts collected in Section~\ref{post_collection}, we counted the operating systems mentioned in discussions involving cross-OS translation, adaptation, or translation needs. The results show that such needs are mainly concentrated on six operating systems: FreeRTOS (18 mentions), Zephyr (15 mentions), RT-Thread (11 mentions), NuttX (9 mentions), Unix (13 mentions), and HarmonyOS (7 mentions). Based on this distribution, we selected these six operating systems for the cross-OS translation task and used them to guide the construction of the subsequent parallel corpus.

\textbf{Pairwise Code Mapping.} For the cross-OS translation task defined in Equation~\ref{eq:os_translation}, we constructed task data by mining isomorphic implementations across the six selected operating systems. These implementations were mainly collected from embedded applications with multi-OS abstraction layers and adapted versions of third-party libraries in official OS repositories. We first identified OS-agnostic abstraction interfaces and then located their corresponding concrete implementations in each operating system, based on which we paired implementations across different operating systems. To ensure mapping quality, two reviewers examined more than 650 aligned function pairs and checked their semantic equivalence, interface consistency, and control-flow compatibility, correcting or removing more than 30 pairs with questionable equivalence. This process finally resulted in 610 translation pairs.

\begin{figure}[!t]
\centering
\includegraphics[width=\columnwidth]{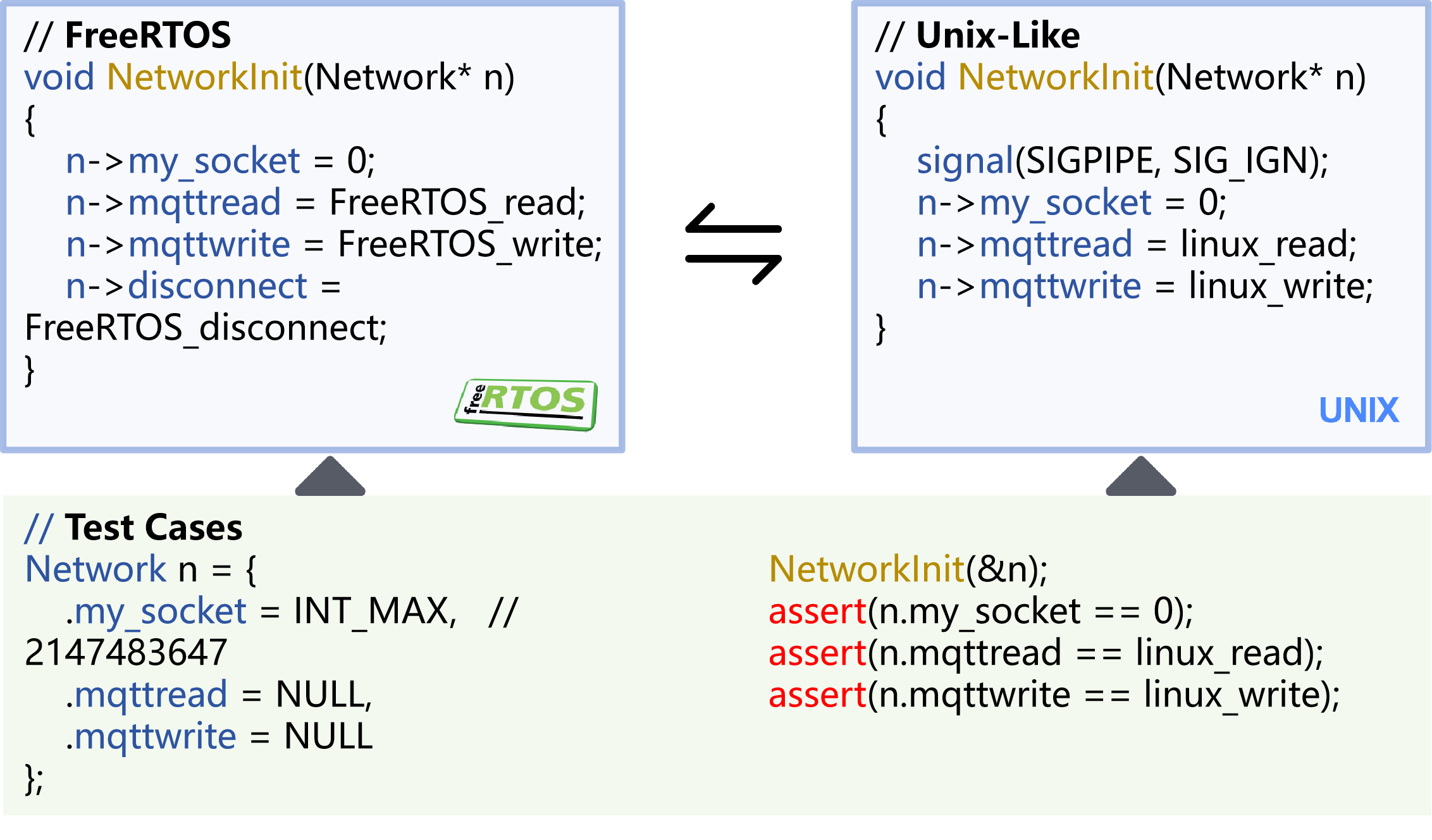}
\caption{Cross-OS data example}
\label{fig:cross_os}
\end{figure}

\begin{figure}[!t]
\centering
\includegraphics[width=\columnwidth]{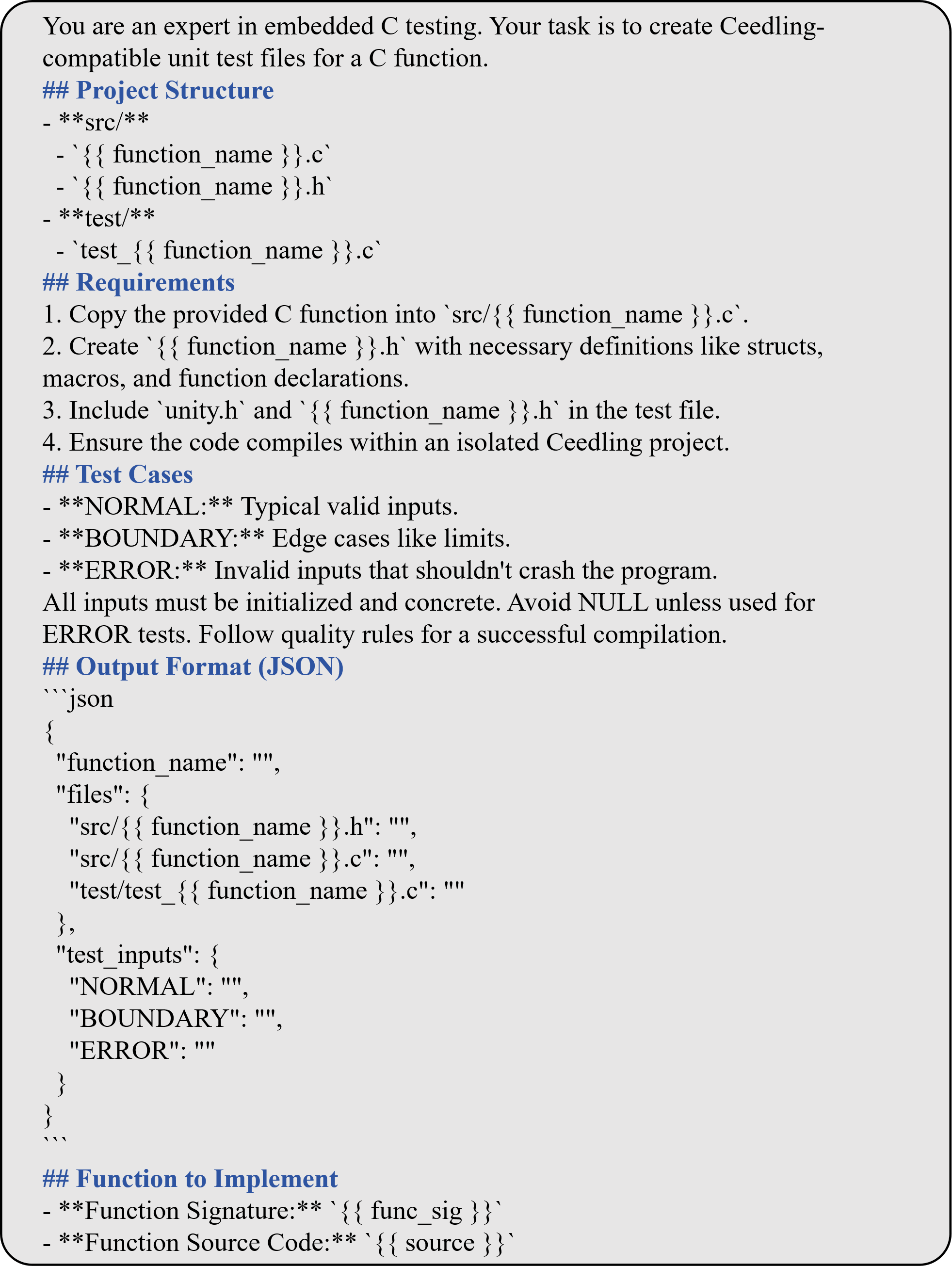}
\caption{Test Generation Prompt}
\label{fig:generate_test_prompt}
\end{figure}

\textbf{Test Case Generation.} Since translation quality depends on semantic equivalence rather than syntactic similarity, we attached executable test cases to the translation pairs for dynamic checking. For each pair, we used DeepSeek-V3.2 to generate test inputs and then executed both the source and translated implementations in a Ceedling~\cite{ceedling_build_system} and Unity~\cite{unity_test_framework} environment. The prompt used for test case generation is shown in Fig.~\ref{fig:generate_test_prompt}. A translated function is considered correct only if it matches the source implementation on all test cases. On average, each pair is associated with three test cases. Figure~\ref{fig:cross_os} presents a representative example, where the \texttt{NetworkInit} function is implemented differently on FreeRTOS and Unix-like systems while preserving the same abstract behavior and can be validated through unit tests.

\begin{table}[!t]
\caption{Composition of RealEmbedBench}
\label{tab:tasks}
\centering
\small
\begin{tabularx}{\columnwidth}{@{}l c X@{}}
\toprule
\textbf{Task} & \textbf{Size} & \textbf{Statistics} \\
\midrule
Code Gen. / Completion
& 1,426 / 3,026
& 56 LOC; 4.93 calls; 3.91 globals; 3.98 types \\
\midrule
State Machine
& 80
& 7.73 states; 14.05 transitions \\
\midrule
Cross-OS Translation
& 610
& 6 OSs; 26 directions \\
\bottomrule
\end{tabularx}
\end{table}

\begin{table*}[!t]
\centering
\caption{Cross-OS Translation Task Distribution}
\label{tab:adaptation-distribution}
\setlength{\tabcolsep}{12pt} 
\scriptsize
\begin{tabular}{lcccccc}
\toprule
\textbf{OS Type} & \textbf{FreeRTOS} & \textbf{RT-Thread} & \textbf{Unix-like} & \textbf{Zephyr} & \textbf{NuttX} & \textbf{Harmony} \\
\midrule
\textbf{FreeRTOS}  & - & 49 & 48 & 24 & 20 & 18 \\
\textbf{RT-Thread} & 49 & - & 37 & 21 & 18 & 1 \\
\textbf{Unix-like} & 48 & 37 & - & 23 & 22 & 4 \\
\textbf{Zephyr}    & 24 & 21 & 23 & - & 20 & 0 \\
\textbf{NuttX}     & 20 & 18 & 22 & 20 & - & 0 \\
\textbf{Harmony}   & 18 & 1 & 4 & 0 & 0 & - \\
\bottomrule
\end{tabular}
\begin{flushleft}
\footnotesize{Note: values indicate the number of code adaptation tasks migrating from source to target OS.}
\end{flushleft}
\end{table*}

\subsection{Benchmark Statistics}

Table~\ref{tab:tasks} summarizes the composition of RealEmbedBench. It includes 1,426 code generation samples and 3,026 code completion samples, alongside 80 state machine samples and 610 cross-OS translation samples. For code generation \& completion, the average function length is 56 LOC, with 4.93 call dependencies, 3.91 global-variable dependencies, and 3.98 type dependencies per sample. State machine samples contain 7.73 states and 14.05 transitions on average, and the cross-OS translation task covers 6 embedded operating systems and 26 directed translation pairs, as detailed in Table~\ref{tab:adaptation-distribution}.

\section{EVALUATION SETUP}
\label{sec:evaluation}
In this section, we introduce the overall experimental setup, including the research questions, the selected evaluation models, the evaluation metrics for different tasks, and the implementation details.

\subsection{Studied Models}
To comprehensively evaluate the coding capabilities and domain understanding required by current large language models in embedded scenarios, we selected six representative instruction-tuned LLMs. Specifically, we employ five large-scale general-purpose models, including DeepSeek-V3.2 (671B)~\cite{deepseek2025deepseek}, Qwen3-235B-A22B-Instruct~\cite{yang2025qwen3}, GLM-4.7 (355B)~\cite{zhipu2025glm47}, GPT-5.2~\cite{openai2025gpt52}, and Gemini-3.1-Pro~\cite{google2025gemini31pro}. In addition, we include StarCoder (15.5B)~\cite{bigcode2024starcoder215b}, a code-specialized LLM, to examine whether domain-specific pretraining brings extra advantages in embedded code tasks. These models cover both open-source and closed-source categories and are representative recent models available at the time of evaluation.

\subsection{Evaluation Metrics}
\subsubsection*{Code Generation and Completion Tasks}
For code generation and block-level code completion tasks, we adopt \textbf{CodeBLEU}~\cite{ren2020codebleu} as the evaluation metric. For token-level code completion, since CodeBLEU is not suitable for evaluating very short local predictions, we use Exact Match to directly determine whether the generated token exactly matches the reference token.

\subsubsection{Natural Language to PlantUML}
To evaluate the quality of generated PlantUML state machine diagrams, we introduce three automatic metrics: Syntactic Validity, which checks whether the generated diagram conforms to PlantUML syntax; State Coverage, which measures the proportion of correctly covered reference states; and Transition Coverage, which measures the proportion of correctly covered reference transitions. Detailed definitions are shown in Table~\ref{tab:evaluation_metrics}.

\subsubsection{Natural Language to Code \& PlantUML to Code}
For subtasks where the output target is code, we employ \textbf{CodeBLEU}~\cite{ren2020codebleu}. To avoid the influence of environment setup and boilerplate code, we only evaluate the main loop logic and core state switching modules.

\subsubsection{Cross-OS Translation Task}
For cross-OS translation, we use \textbf{pass@1} as the core metric. Each model generates one candidate program for each test case, and the result is considered successful only if the generated code compiles, executes, and passes all test cases.

\begin{table}[htbp]
\centering
\caption{Evaluation Metrics Description}
\label{tab:evaluation_metrics}
\scriptsize
\renewcommand{\arraystretch}{1.4} 
\newcolumntype{M}{>{\raggedright\arraybackslash\hsize=0.2\hsize}X}
\newcolumntype{D}{>{\raggedright\arraybackslash\hsize=0.5\hsize}X}
\newcolumntype{F}{>{\centering\arraybackslash\hsize=0.3\hsize}X}
\begin{tabularx}{\columnwidth}{@{} M D F @{}}
\toprule
\textbf{Metric} & \textbf{Description} & \textbf{Formula} \\
\midrule
Syntactic Validity & Evaluate the syntactic correctness of PlantUML diagrams & Whether the PlantUML passes compilation \\
\midrule
State Machine State Coverage & Measure the proportion of correctly covered reference states & $|P_S \cap R_S|/|R_S|$ \\
\midrule
State Transition Coverage & Measure the proportion of correctly covered reference transitions & $|P_T \cap R_T|/|R_T|$ \\
\bottomrule
\end{tabularx}

\begin{flushleft}
\footnotesize
Note: $P_S$ = PredictedStates, $R_S$ = ReferenceStates, $P_T$ = PredictedTransitions, $R_T$ = ReferenceTransitions.
\end{flushleft}
\end{table}

\subsection{Implementation Details}
For the locally deployed StarCoder model, inference is performed on a single A100 GPU using the vLLM framework, while the remaining models are accessed through their official APIs. For all models, the temperature is set to 0.6, top-$p$ is set to 0.95, and the maximum context window is limited to 8192 tokens. Each task uses a task-specific prompt template, which will be released in our code repository. For code generation and completion tasks, CodeBLEU is computed with four equally weighted sub-metrics: \texttt{ngram\_match}, \texttt{weighted\_ngram\_match}, \texttt{syntax\_match}, and \texttt{dataflow\_match}.

\section{Results}

In this section, we present and discuss the results related to our evaluation tasks. For all tasks, we follow the experiment setup described in Section~\ref{sec:evaluation}.

\subsection{RQ1: How do different large language models perform on code generation and completion tasks?}

\textbf{Observation 1:} \textbf{Different large language models exhibit clear performance differences on embedded code generation and completion tasks.} As shown in Table~\ref{tab:model_performance_abbr}, the evaluated models achieve notably different results on both CodeGen and CodeComp, indicating varying capabilities in handling embedded programming tasks. For example, CodeGen scores range from 0.4823 to 0.7076, while CodeComp scores range from 0.4756 to 0.6915. This suggests that model choice has a substantial impact on performance in embedded code synthesis and completion.

\textbf{Observation 2:} \textbf{\texttt{gemini-3.1-pro} performs best overall on both code generation and code completion tasks.} According to Table~\ref{tab:model_performance_abbr}, \texttt{gemini-3.1-pro} achieves the highest score on CodeGen (0.7076) and also ranks first on CodeComp (0.6915), outperforming all other compared models. This result indicates that \texttt{gemini-3.1-pro} has the strongest overall capability for embedded code-related tasks among the evaluated models.

\subsection{RQ2: How do different large language models perform on state machine understanding and implementation tasks?}

\textbf{Observation 3:} \textbf{No single large language model achieves the best performance across all state-machine-related tasks.} As shown in Table~\ref{tab:model_performance_abbr}, the best results are distributed across different models rather than concentrated in one model. While some models perform better on structural understanding tasks, others achieve stronger results on code synthesis tasks. This suggests that state machine understanding and implementation involve multiple capabilities, and no model consistently dominates all aspects.

\textbf{Observation 4:} \textbf{\texttt{gemini-3.1-pro} performs best on three state-machine-related tasks, namely R2P-C, P2C, and R2C.} According to Table~\ref{tab:model_performance_abbr}, \texttt{gemini-3.1-pro} achieves the highest score on PlantUML compilation rate (R2P-C, 0.9601), PlantUML-to-Code (P2C, 0.4177), and Requirement-to-Code (R2C, 0.4379). These results indicate that \texttt{gemini-3.1-pro} shows particular strength in generating valid state machine representations and translating state-machine-related inputs into executable code.

\textbf{Observation 5:} \textbf{\texttt{deepseek-v3.2} performs best on the R2P-N task.} As shown in Table~\ref{tab:model_performance_abbr}, \texttt{deepseek-v3.2} obtains the highest score on Req2PlantUML node coverage (R2P-N, 0.7598), outperforming all other models on this metric. This suggests that it is particularly effective at identifying and representing key states from natural language requirements.

\subsection{RQ3: How do different large language models perform on cross-OS translation tasks?}

\textbf{Observation 6:} \textbf{\texttt{glm-4.7} achieves the highest overall average performance on cross-OS translation tasks.} As shown in Table~\ref{tab:model_performance_abbr}, \texttt{glm-4.7} obtains the best average CodeTrans score of 0.7833, outperforming all other evaluated models. It is followed by \texttt{deepseek-v3.2} (0.7529) and \texttt{gemini-3.1-pro} (0.7481). This result indicates that \texttt{glm-4.7} is the strongest model overall for cross-OS code translation in our benchmark.

\textbf{Observation 7:} \textbf{Different models show substantial variation across conversion categories, and translation performance is clearly direction-dependent.} Table~\ref{tab:conversion_metrics_split} shows that the best-performing model varies considerably across source--target OS pairs. For example, \texttt{gemini-3.1-pro} performs best on \texttt{RT-Thread to FreeRTOS} (0.788), \texttt{FreeRTOS to RT-Thread} (0.786), and \texttt{Zephyr to RT-Thread} (0.917), whereas \texttt{glm-4.7} achieves the best results on \texttt{NuttX to RT-Thread} (1.000) and \texttt{Unix to RT-Thread} (0.903), and \texttt{deepseek-v3.2} performs best on \texttt{FreeRTOS to Zephyr} (0.824). In addition, the results reveal clear asymmetry between opposite directions. For instance, \texttt{RT-Thread to Harmony} reaches 1.000 for all models, while \texttt{Harmony to RT-Thread} is much more difficult, with most models scoring 0.000. This suggests that cross-OS translation difficulty depends not only on the OS pair itself but also on the translation direction.

\textbf{Observation 8:} \textbf{\texttt{Harmony to RT-Thread} is the most challenging conversion direction overall, with the lowest average pass rate across models.} According to Table~\ref{tab:conversion_metrics_split}, five out of six models score 0.000 on this direction, and only \texttt{glm-4.7} reaches 1.000, resulting in an average pass rate of only 0.1667. Compared with other conversion directions, this is the lowest average result in the benchmark, indicating that translating from Harmony to RT-Thread remains particularly difficult for current large language models. The comparatively strong performance of \texttt{glm-4.7} on this direction may be related to more sufficient exposure to HarmonyOS-related training data, which could give it an advantage in handling Harmony-specific APIs and programming patterns.

\begin{table*}[htbp]
\centering
\caption{Performance comparison of models on different tasks (maximum values in bold red).}
\label{tab:model_performance_abbr}
\small
\setlength{\tabcolsep}{4pt}
\renewcommand{\arraystretch}{1.15}

\resizebox{\textwidth}{!}{%
\begin{tabular}{l | c c | c c c c c | c}
\toprule
\multicolumn{1}{l|}{Task} 
& \multicolumn{2}{c|}{Code Gen./Comp.} 
& \multicolumn{5}{c|}{State Machine} 
& \multicolumn{1}{c}{Code Trans.} \\
\midrule
Model & CodeGen & CodeComp & R2P-C & R2P-N & R2P-E & P2C & R2C & CodeTrans \\
\midrule
deepseek-v3.2            & 0.5388 & 0.5738 & 0.9567 & \textbf{\textcolor{red}{0.7598}} & 0.6467 & 0.3848 & 0.3809 & 0.7529 \\
qwen3-235b-a22b-instruct & 0.4916 & 0.5211 & 0.9523 & 0.7508 & 0.5601 & 0.4100 & 0.4148 & 0.7371 \\
glm-4.7                  & 0.4901 & 0.4804 & 0.9589 & 0.7334 & 0.6766 & 0.3902 & 0.3934 & \textbf{\textcolor{red}{0.7833}} \\
gemini-3.1-pro           & \textbf{\textcolor{red}{0.7076}} & \textbf{\textcolor{red}{0.6915}} & \textbf{\textcolor{red}{0.9601}} & 0.7183 & 0.6566 & \textbf{\textcolor{red}{0.4177}} & \textbf{\textcolor{red}{0.4379}} & 0.7481 \\
gpt-5.2                  & 0.5095 & 0.6812 & 0.9545 & 0.7493 & \textbf{\textcolor{red}{0.6914}} & 0.3750 & 0.3706 & 0.6982 \\
starcoder                & 0.3823 & 0.4756 & 0.3552 & 0.3134 & 0.1532 & 0.1689 & 0.2612 & 0.1404 \\
\bottomrule
\end{tabular}%
}

\vspace{3pt}
\begin{flushleft}
\footnotesize\textit{Note:} CodeGen = Code Generation, CodeComp = Code Completion, R2P-C = PlantUML Compilation Rate, R2P-N = Req2PlantUML (Node), R2P-E = Req2PlantUML (Edge), P2C = PlantUML2Code, R2C = Req2Code, CodeTrans = Code Translation.
\end{flushleft}
\end{table*}

\begin{table*}[htbp]
\centering
\caption{Performance metrics (pass rate) for different model conversion tasks}
\label{tab:conversion_metrics_split}
\small
\setlength{\tabcolsep}{3pt}
\renewcommand{\arraystretch}{1.2}

\begin{minipage}{\linewidth}
\centering
\textbf{(a) Conversion pairs 1--7}\par\smallskip
\begin{tabular}{l *{7}{c}}
\toprule
Model & \makecell{RT-Thread\\to FreeRTOS} & \makecell{RT-Thread\\to Harmony} & \makecell{RT-Thread\\to NuttX} & \makecell{RT-Thread\\to Unix} & \makecell{RT-Thread\\to Zephyr} & \makecell{FreeRTOS\\to RT-Thread} & \makecell{FreeRTOS\\to Harmony} \\
\midrule
deepseek-v3.2  & 0.643 & \cellcolor{blue!25}1.000 & 0.812 & 0.650 & 0.917 & 0.697 & 0.467 \\
glm-4.7        & 0.536 & \cellcolor{blue!25}1.000 & 0.778 & 0.609 & 0.923 & 0.700 & 0.538 \\
qwen3-235b     & 0.667 & \cellcolor{blue!25}1.000 & 0.765 & 0.706 & 0.909 & 0.727 & 0.333 \\
gpt-5.2        & 0.600 & \cellcolor{blue!25}1.000 & 0.722 & \cellcolor{blue!25}0.750 & 0.923 & 0.667 & 0.429 \\
gemini-3.1-pro & \cellcolor{blue!25}0.788 & \cellcolor{blue!25}1.000 & 0.667 & 0.688 & \cellcolor{blue!25}0.933 & \cellcolor{blue!25}0.786 & \cellcolor{blue!25}0.722 \\
starcoder      & 0 & \cellcolor{blue!25}1.000 & 0 & 0.215 & 0.247 & 0 & 0.191 \\
\bottomrule
\end{tabular}
\end{minipage}

\par\vspace{1em}

\begin{minipage}{\linewidth}
\centering
\textbf{(b) Conversion pairs 8--14}\par\smallskip
\begin{tabular}{l *{7}{c}}
\toprule
Model & \makecell{FreeRTOS\\to NuttX} & \makecell{FreeRTOS\\to Unix} & \makecell{FreeRTOS\\to Zephyr} & \makecell{Harmony\\to RT-Thread} & \makecell{Harmony\\to FreeRTOS} & \makecell{Harmony\\to Unix} & \makecell{NuttX\\to RT-Thread} \\
\midrule
deepseek-v3.2  & 0.938 & 0.688 & \cellcolor{blue!25}0.824 & 0.000 & 0.500 & \cellcolor{blue!25}1.000 & 0.818 \\
glm-4.7        & 0.941 & 0.594 & 0.722 & \cellcolor{blue!25}1.000 & 0.500 & \cellcolor{blue!25}1.000 & \cellcolor{blue!25}1.000 \\
qwen3-235b     & 0.824 & 0.615 & 0.786 & 0.000 & 0.556 & \cellcolor{blue!25}1.000 & 0.818 \\
gpt-5.2        & 0.588 & 0.629 & 0.737 & 0.000 & \cellcolor{blue!25}0.750 & \cellcolor{blue!25}1.000 & 0.917 \\
gemini-3.1-pro & 0.778 & \cellcolor{blue!25}0.721 & 0.810 & 0.000 & 0.688 & \cellcolor{blue!25}1.000 & 0.917 \\
starcoder      & 0 & 0.142 & 0.189 & 0 & 0.203 & 0 & 0.137 \\
\bottomrule
\end{tabular}
\end{minipage}

\par\vspace{1em}

\begin{minipage}{\linewidth}
\centering
\textbf{(c) Conversion pairs 15--21}\par\smallskip
\begin{tabular}{l *{7}{c}}
\toprule
Model & \makecell{NuttX\\to FreeRTOS} & \makecell{NuttX\\to Unix} & \makecell{NuttX\\to Zephyr} & \makecell{Unix\\to RT-Thread} & \makecell{Unix\\to FreeRTOS} & \makecell{Unix\\to Harmony} & \makecell{Unix\\to NuttX} \\
\midrule
deepseek-v3.2  & 0.643 & 0.692 & \cellcolor{blue!25}1.000 & 0.897 & 0.625 & \cellcolor{blue!25}1.000 & 0.889 \\
glm-4.7        & 0.700 & 0.615 & \cellcolor{blue!25}1.000 & \cellcolor{blue!25}0.903 & 0.625 & \cellcolor{blue!25}1.000 & 0.895 \\
qwen3-235b     & \cellcolor{blue!25}0.800 & 0.545 & 0.933 & 0.808 & \cellcolor{blue!25}0.737 & \cellcolor{blue!25}1.000 & 0.905 \\
gpt-5.2        & 0.538 & 0.500 & 0.944 & 0.880 & 0.724 & 0.500 & 0.700 \\
gemini-3.1-pro & 0.583 & \cellcolor{blue!25}0.706 & 0.941 & 0.839 & 0.667 & 0.750 & 0.864 \\
starcoder      & 0 & 0.174 & 0 & 0.198 & 0.251 & 0 & 0.142 \\
\bottomrule
\end{tabular}
\end{minipage}

\par\vspace{1em}

\begin{minipage}{\linewidth}
\centering
\textbf{(d) Conversion pairs 22--26}\par\smallskip
\begin{tabular}{l *{5}{c}}
\toprule
Model & \makecell{Unix\\to Zephyr} & \makecell{Zephyr\\to RT-Thread} & \makecell{Zephyr\\to FreeRTOS} & \makecell{Zephyr\\to NuttX} & \makecell{Zephyr\\to Unix} \\
\midrule
deepseek-v3.2  & 0.857 & 0.786 & \cellcolor{blue!25}0.733 & \cellcolor{blue!25}0.812 & 0.688 \\
glm-4.7        & 0.800 & 0.750 & 0.688 & \cellcolor{blue!25}0.812 & 0.737 \\
qwen3-235b     & 0.833 & 0.765 & \cellcolor{blue!25}0.733 & \cellcolor{blue!25}0.812 & 0.588 \\
gpt-5.2        & \cellcolor{blue!25}0.900 & 0.818 & 0.562 & 0.765 & 0.611 \\
gemini-3.1-pro & \cellcolor{blue!25}0.900 & \cellcolor{blue!25}0.917 & 0.684 & \cellcolor{blue!25}0.812 & \cellcolor{blue!25}0.750 \\
starcoder      & 0 & 0.158 & 0.192 & 0 & 0.211 \\
\bottomrule
\end{tabular}
\end{minipage}

\vspace{2mm}
\begin{flushleft}
\footnotesize
Note: Each cell shows pass rate. Pass rate is the proportion of executable code that passes three test cases. Cells with the highest value in each column are highlighted in light blue.
\end{flushleft}
\end{table*}

\section{DISCUSSION}
In this section, we discuss the results on the three task categories in RealEmbedBench. For embedded software code generation, we analyze the effect of different context configurations through ablation experiments and inspect representative cases to understand the role of each context type. For state machine modeling and implementation, we examine incorrect cases to identify common error patterns. For cross-OS translation, we analyze unsuccessful translations to summarize major semantic mismatch patterns.

\subsection{Discussion on Embedded Software Code Generation Tasks}

To investigate how different context types influence embedded software code generation, we compare seven context configurations across six models. Our focus is on two main aspects: the overall contribution of contextual information and identifying which context type is the most important, supported by representative examples and ablation experiments for analysis.

\begin{figure}[!t]
\centering
\includegraphics[width=\columnwidth]{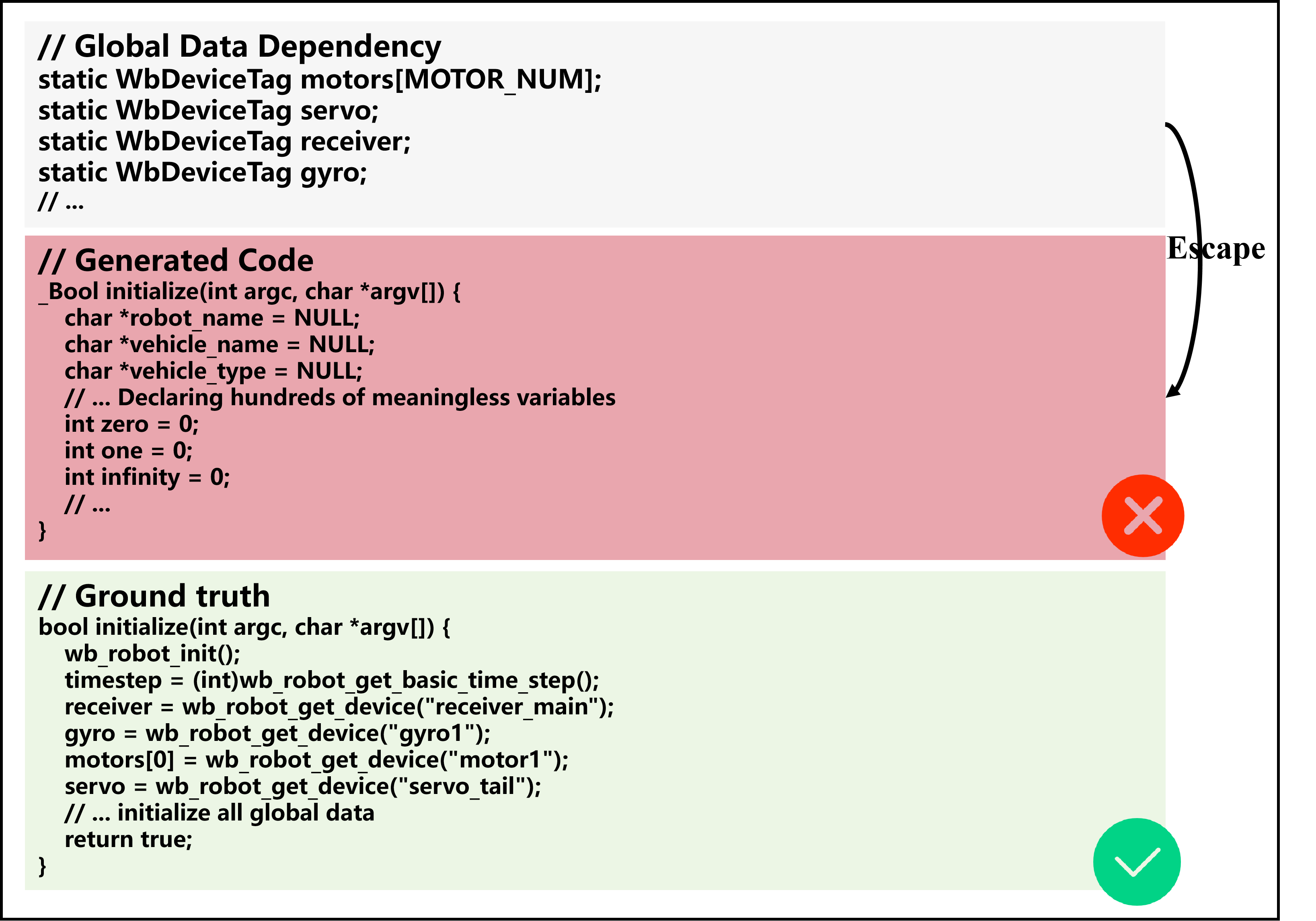}
\caption{Escape Global Data Example} 
\label{fig:global_data_esc}
\end{figure}

\begin{figure}[!t]
\centering
\includegraphics[width=\columnwidth]{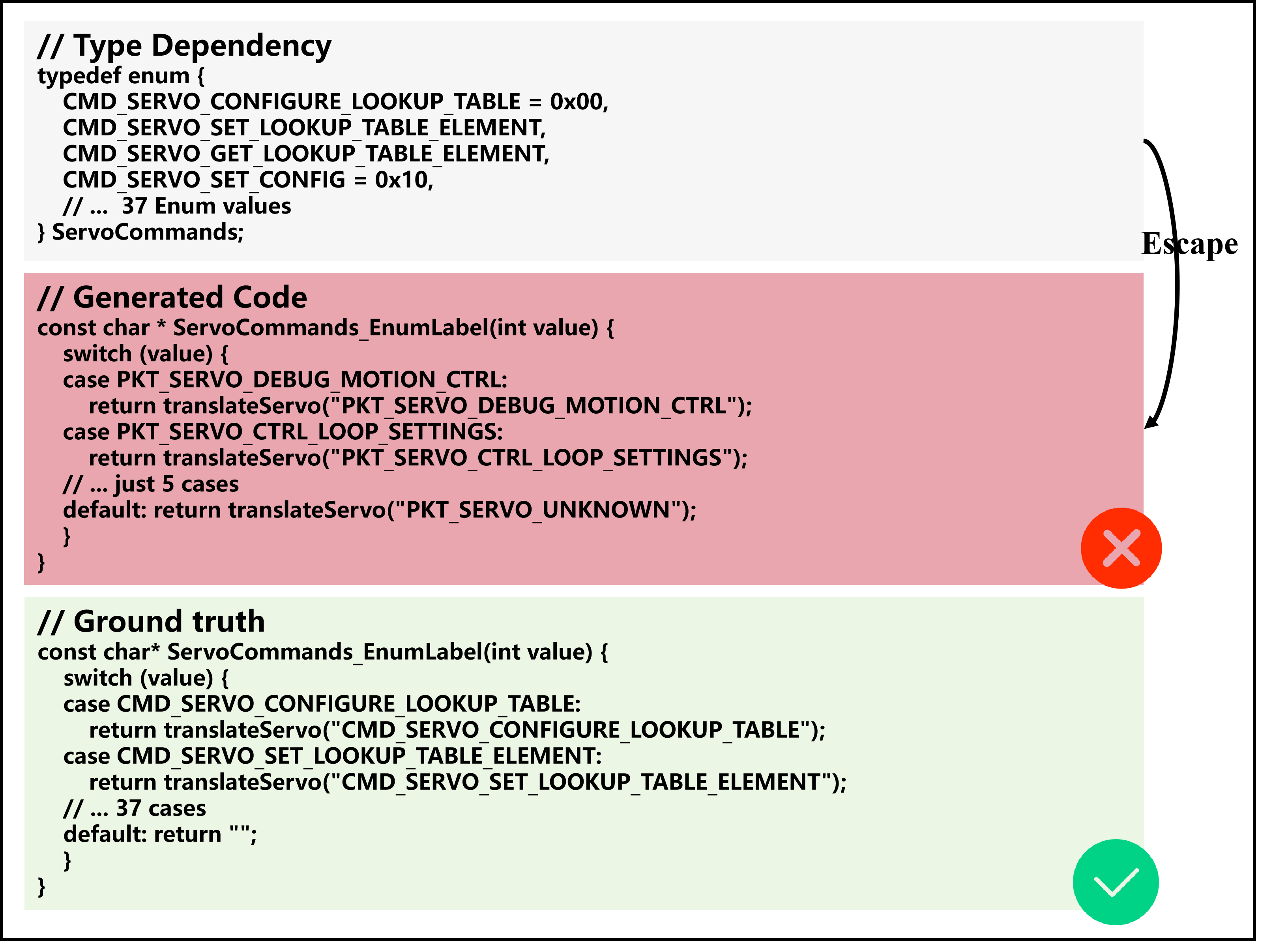}
\caption{Escape Type Dependency Example} 
\label{fig:type_dependency_esc}
\end{figure}

\textbf{Context has a substantial impact on model performance in embedded software code generation.}
As shown in Table~\ref{tab:context_escape_analysis}, all six models exhibit their largest performance drop under the \texttt{NoCtx} setting. For example, gemini-3.1-pro decreases from 0.7076 to 0.4149, deepseek-v3.2 from 0.5388 to 0.3260, and gpt-5.2 from 0.5095 to 0.2466. This consistent decline indicates that context is essential for embedded software generation. The reason is that embedded code is usually strongly constrained by project-specific states, interfaces, and data structures, so generating from the target span alone is often insufficient. Even context types with relatively smaller average impact, such as global data and type definitions, still provide important semantic support. As illustrated in Figure~\ref{fig:global_data_esc} and Figure~\ref{fig:type_dependency_esc}, removing these contexts may not completely prevent generation, but it makes the output more likely to deviate from the actual project semantics.

\begin{figure}[!t]
\centering
\includegraphics[width=\columnwidth]{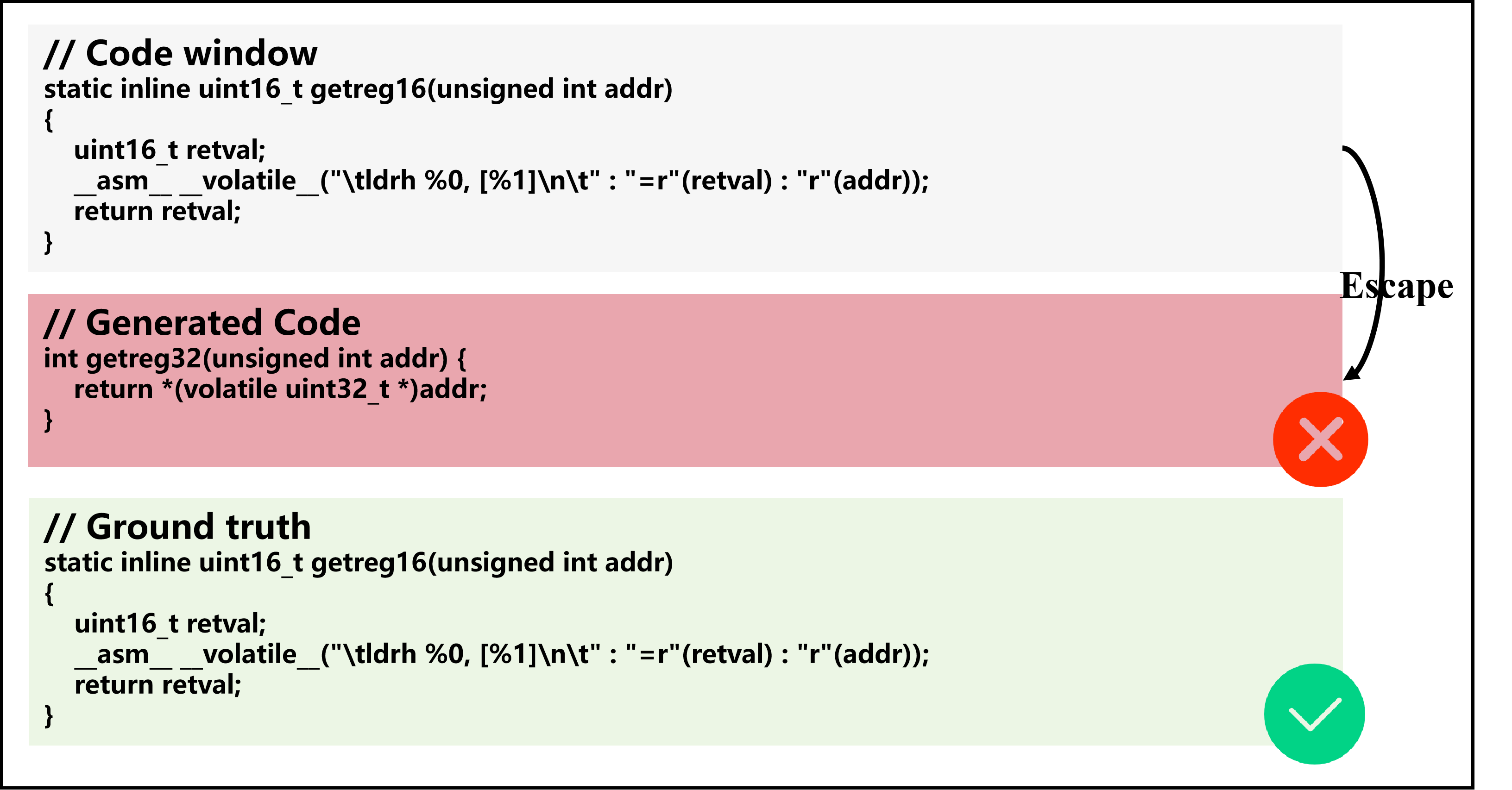}
\caption{Escape Code Window Example} 
\label{fig:code_window_esc}
\end{figure}

\textbf{Among all context types, the local code window has the strongest effect on model performance.}
As shown in Table~\ref{tab:context_escape_analysis}, \texttt{CodeWinEsc} leads to the largest and most consistent drop among all single-context removal settings across the six models, with relative decreases ranging from 13.8\% to 16.9\%. This suggests that nearby code is the most critical context for embedded software generation. A likely reason is that embedded software usually follows strict coding conventions and regular file-level implementation patterns, so the target code often needs to align closely with surrounding control flow, helper usage, and operation order. Without the local code window, the model can still generate plausible code, but it is more likely to break the style and logic consistency of the current file. This is also illustrated in Figure~\ref{fig:code_window_esc}, where the generated code remains locally plausible, but deviates from the surrounding implementation pattern when the local code window is unavailable.

\begin{table*}[!t]
\centering
\caption{Code generation performance of different models under seven context configurations.}
\label{tab:context_escape_analysis}
\small
\renewcommand{\arraystretch}{1.3}
\begin{tabular}{lccccccc}
\toprule
Model & Full & ImpEsc & CodeWinEsc & CallDepEsc & GlobDataEsc & TypeDepEsc & NoCtx \\
\midrule
deepseek-v3.2            & 0.5388 & 0.5360 & 0.4642 & 0.5063 & 0.5273 & 0.5236 & 0.3260 \\
qwen3-235b-a22b-instruct & 0.4916 & 0.4893 & 0.4148 & 0.4555 & 0.4728 & 0.4783 & 0.2451 \\
glm-4.7                  & 0.4901 & 0.4855 & 0.4204 & 0.4608 & 0.4744 & 0.4628 & 0.2938 \\
gemini-3.1-pro           & \textbf{\textcolor{red}{0.7076}} & \textbf{\textcolor{red}{0.6907}} & \textbf{\textcolor{red}{0.5929}} & \textbf{\textcolor{red}{0.6712}} & \textbf{\textcolor{red}{0.6739}} & \textbf{\textcolor{red}{0.6857}} & \textbf{\textcolor{red}{0.4149}} \\
gpt-5.2                  & 0.5095 & 0.4773 & 0.4235 & 0.4522 & 0.4911 & 0.4673 & 0.2466 \\
starcoder                & 0.3823 & 0.3775 & 0.3082 & 0.3699 & 0.3654 & 0.3605 & 0.3676 \\
\bottomrule
\end{tabular}

\vspace{3pt}
\begin{flushleft} 
\footnotesize \textit{Note:} Full = full context; ImpEsc = import dependency escaped; CodeWinEsc = in-file code window only removed; CallDepEsc = call dependency escaped; GlobDataEsc = global data dependency escaped; TypeDepEsc = type definition dependency escaped; NoCtx = no context provided.
\end{flushleft}
\end{table*}

\subsection{Discussion on State Machine Modeling and Implementation Tasks}

\begin{table}[!t]
\centering
\caption{Summary of major error patterns in state machine tasks}
\label{tab:stm_error_patterns}
\small
\setlength{\tabcolsep}{4pt}
\renewcommand{\arraystretch}{1.15}
\begin{tabularx}{\columnwidth}{
>{\raggedright\arraybackslash}X
>{\centering\arraybackslash}p{0.18\columnwidth}}
\toprule
\textbf{Error Pattern} & \textbf{Freq.} \\
\midrule
API usage mismatch & 38\% \\
Misplaced condition evaluation & 32\% \\
Transition logic simplification & 22\% \\
State structure over-complication & 14\% \\
\bottomrule
\end{tabularx}

\vspace{2pt}
\begin{flushleft}
\footnotesize
\textit{Note:} Freq. denotes the proportion of 50 sampled incorrect cases in which each error pattern appears.
\end{flushleft}
\end{table}

To identify the main challenges in state machine modeling and implementation, we analyze 50 incorrect cases and summarize the major error patterns in Table~\ref{tab:stm_error_patterns}. The following discussion focuses on the typical ways in which model outputs diverge from the reference implementation.

\textbf{API hallucination.} As shown in Table~\ref{tab:stm_error_patterns}, API hallucination is the most prevalent error pattern, accounting for 38\% of the analyzed cases. This type of error occurs when the model fails to follow the actual interface conventions of the target project and instead invents plausible but unsupported APIs or helper abstractions. In the example shown in the figure, the reference implementation determines whether WiFi is enabled by directly accessing the member variable wifi\_enabled\_, whereas the generated code replaces this access with a call to this->is\_wifi\_enabled(). Although this generated API appears semantically reasonable, it is not part of the target project’s actual interface design and therefore constitutes a typical case of API hallucination. Moreover, this substitution also alters the structure of the original conditional logic and introduces additional behavioral discrepancies between the generated output and the reference implementation.

\begin{figure}[!t]
\centering
\includegraphics[width=\columnwidth]{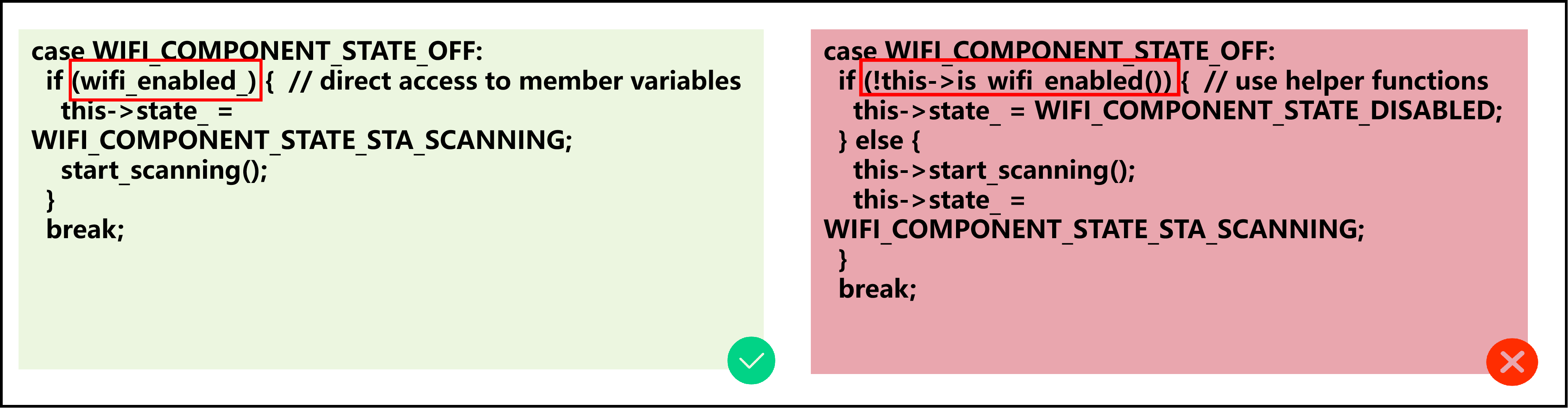}
\caption{State Machine Api Mismatch Example} 
\label{fig:stm_api_mismatch_example}
\end{figure}

\begin{figure}[!t]
\centering
\includegraphics[width=\columnwidth]{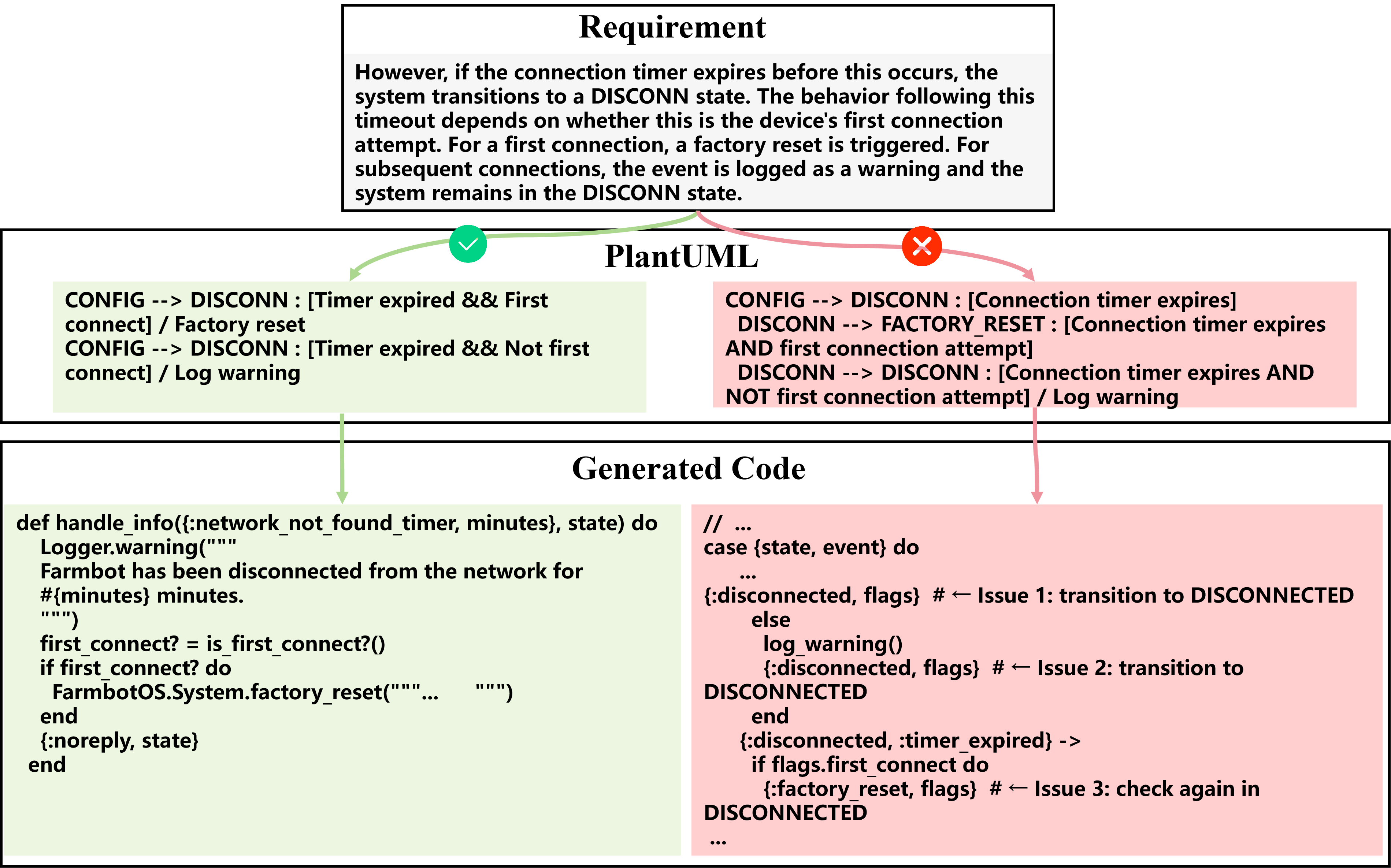}
\caption{State Machine Error Example} 
\label{fig:stm_err_example}
\end{figure}

\textbf{Misplaced condition evaluation.} Misplaced condition evaluation is observed in 32\% of the analyzed cases. This error typically occurs when the model identifies a key condition from the requirement but fails to evaluate it at the correct state or stage of event handling, thereby altering the control semantics of the state machine. As illustrated in Figure~\ref{fig:stm_err_example}, the requirement explicitly specifies that when the connection timer expires, the system should enter the \texttt{DISCONN} state, and that the subsequent handling depends on whether this is the device's first connection: if it is the first connection, a \texttt{Factory reset} should be triggered; otherwise, the system should log a warning and remain in \texttt{DISCONN}. Accordingly, the reference implementation checks both \texttt{timer expired} and \texttt{first connect} at the time the timeout event is handled, encoding them as two guarded transitions from \texttt{CONFIG} to \texttt{DISCONN}, with first-time connection leading to a factory reset and non-first-time connection leading to warning logging. In contrast, although the generated result includes related behavioral elements such as \texttt{DISCONN}, \texttt{Factory reset}, and \texttt{Log warning}, it first transitions to \texttt{DISCONN} in an unconditional or nearly unconditional manner, and only then evaluates \texttt{first\_connect} within the \texttt{DISCONN} state. As a result, a decision that should have been made during timeout handling is postponed until after the system has already entered the disconnected state. Thus, the issue in this example is not that the model omitted the relevant states or conditions, but that it shifted the condition evaluation point to the wrong place, causing the generated state machine to deviate from the required execution order and control semantics.

\begin{figure}[!t]
\centering
\includegraphics[width=\columnwidth]{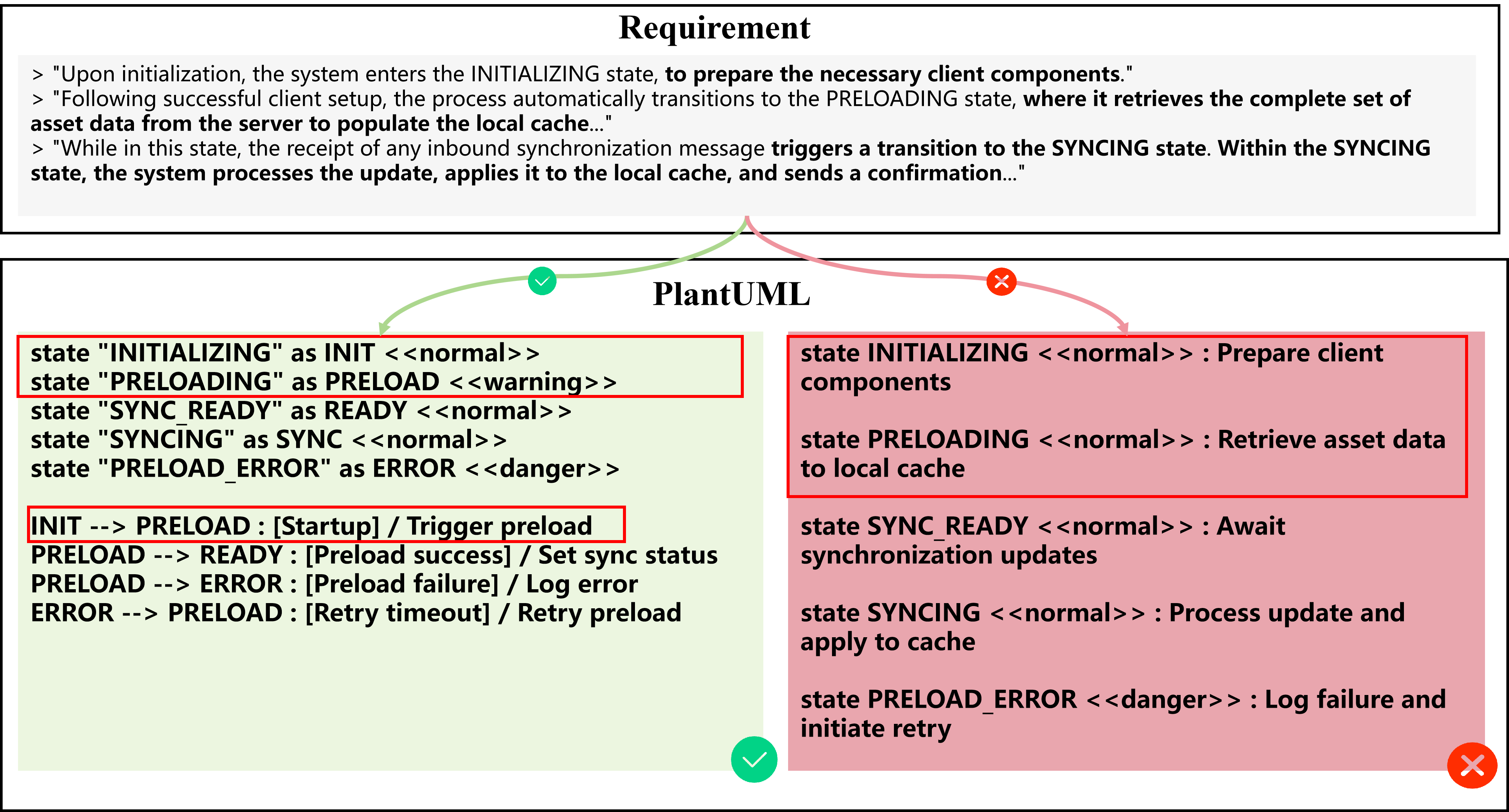}
\caption{Transition Logic Simplification Example} 
\label{fig:stm_miss_trans_example}
\end{figure}

\textbf{Transition logic simplification.} As shown in Table~\ref{tab:stm_error_patterns}, \textit{transition logic simplification} accounts for 22\% of the analyzed cases. This error pattern typically arises when the requirements already specify multiple conditional branches and their corresponding behaviors, but the model fails to preserve these transition conditions in the generated state machine and instead compresses the original fine-grained branching logic into a coarser state progression. As illustrated in Figure~\ref{fig:stm_miss_trans_example}, the requirement not only describes the basic workflow from initialization to preloading and then to synchronization readiness, but also explicitly distinguishes the outcomes of the preloading stage by directing success and failure to different states and by further defining a timeout-triggered retry path after failure. In contrast, although the generated result retains high-level state semantics such as \texttt{INITIALIZING}, \texttt{PRELOADING}, \texttt{SYNC\_READY}, and \texttt{PRELOAD\_ERROR}, it does not faithfully map the conditional transition structure explicitly specified in the requirement. In particular, the distinctions among preload success, preload failure, and failure recovery through retry are not clearly represented as separate guarded transitions, but are instead reduced to a generalized description of process stages. Therefore, the issue in this example does not stem from ambiguity in the requirement itself, but from the model's failure to preserve the branching logic and behavioral constraints that were already explicitly provided. While the generated state machine captures the overall outline of the workflow, it overlooks the conditional branches that are essential to the system's dynamic behavior, thereby reducing fidelity to the semantics of the original state machine.

\begin{figure}[!t]
\centering
\includegraphics[width=\columnwidth]{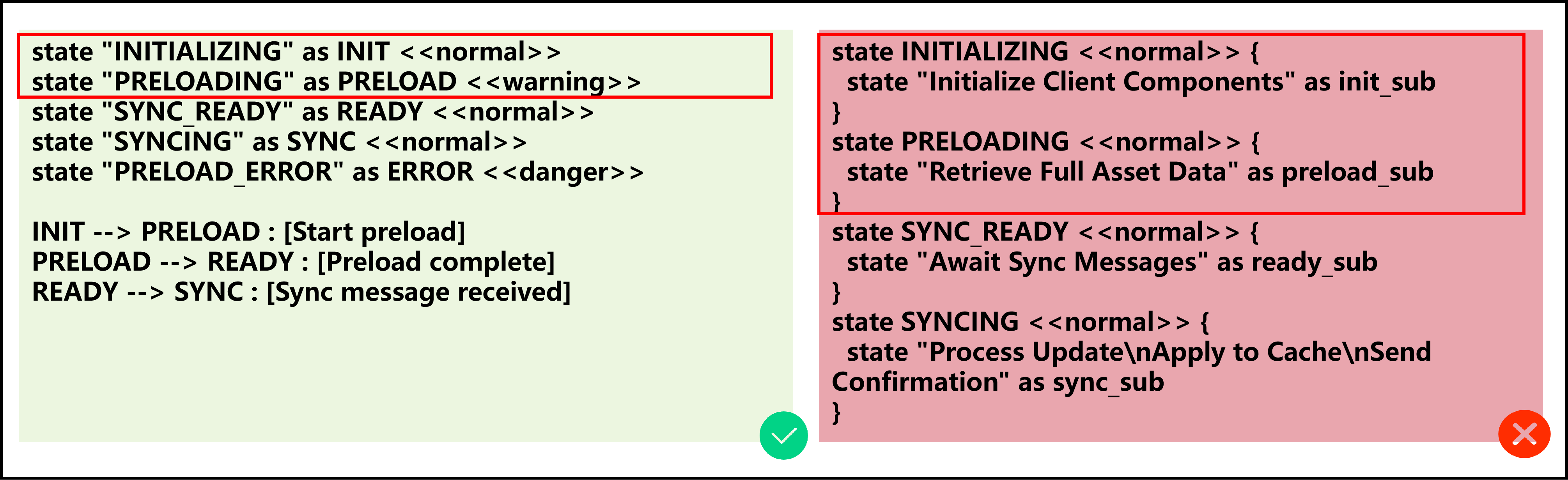}
\caption{State Structure Over-complication Example} 
\label{fig:stm_over_state_example}
\end{figure}

\textbf{State structure over-complication.} As shown in Table~\ref{tab:stm_error_patterns}, \textit{state structure over-complication} accounts for 14\% of the analyzed cases. This error pattern arises when the model fails to preserve the structural granularity of the reference state machine and instead introduces additional hierarchy, nested substates, or unnecessary grouping constructs. As illustrated in Figure~\ref{fig:stm_over_state_example}, the reference state machine follows a relatively flat topology, where states such as \texttt{INITIALIZING}, \texttt{PRELOADING}, \texttt{SYNC\_READY}, and \texttt{SYNCING} are defined at the same hierarchical level and connected through direct transitions. In contrast, the generated output restructures several of these top-level states into composite states and introduces extra internal substates, thereby increasing the hierarchical depth of the model. Although such modifications may appear to improve structural clarity or readability, they alter the original topology of the design and distort the intended level of abstraction.

\subsection{Discussion Cross-OS Translation Tasks}
To investigate the main causes of semantic errors in cross-OS code translation, we analyze 300 unsuccessful translation cases and summarize three major error patterns, as shown in Table~\ref{tab:classic_errors}. 

\begin{table}[!t]
\centering
\caption{Classic Error Types Mentioned in the Document}
\label{tab:classic_errors}
\small
\setlength{\tabcolsep}{5pt}
\renewcommand{\arraystretch}{1.15}
\begin{tabularx}{\columnwidth}{
>{\raggedright\arraybackslash}X
>{\centering\arraybackslash}p{0.24\columnwidth}}
\toprule
\textbf{Error Type} & \textbf{Frequency} \\
\midrule
Omitting Available Functions & 19.21\% \\
Misusing Target Functions & 11.70\% \\
Mistranslating Return Values & 7.80\% \\
\bottomrule
\end{tabularx}
\end{table}

\begin{figure}[!t]
\centering
\includegraphics[width=\columnwidth]{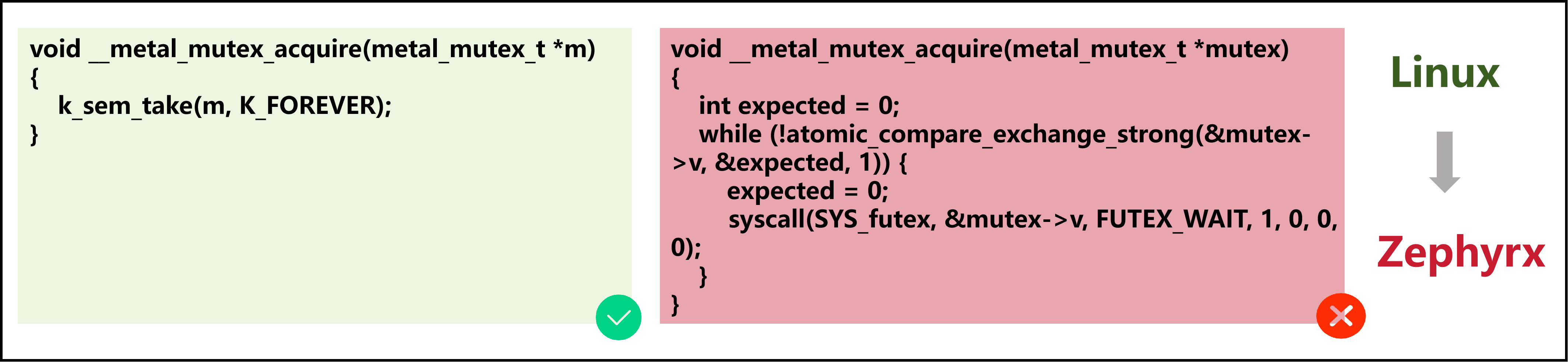}
\caption{Omitting available functions Example}
\label{fig:corss_os_no_use_func}
\end{figure}

\textbf{Omitting available functions.} As shown in Table~\ref{tab:classic_errors}, \textit{omitting available functions} is the most frequent error pattern, accounting for 19.21\% of the analyzed cases. This error occurs when the target OS already provides an appropriate API in the given headers or runtime environment, but the model fails to use it and instead constructs an alternative implementation on its own. Such behavior is problematic because cross-OS translation is expected to map source functionality to native target-side abstractions, rather than re-implementing low-level behavior using unrelated or source-specific techniques. As illustrated in Figure~\ref{fig:corss_os_no_use_func}, in a Linux-to-Zephyr translation example, the correct implementation of \texttt{\_\_metal\_mutex\_acquire} directly invokes Zephyr's built-in synchronization API \texttt{k\_sem\_take(m, K\_FOREVER)}. However, the generated code does not use the available Zephyr function at all. Instead, it introduces a custom lock acquisition routine based on \texttt{atomic\_compare\_exchange\_strong} together with a Linux-style \texttt{futex} \texttt{syscall}. This translation not only ignores the provided target API, but also unnecessarily increases implementation complexity and breaks the expected OS-specific mapping strategy.

\begin{figure}[!t]
\centering
\includegraphics[width=\columnwidth]{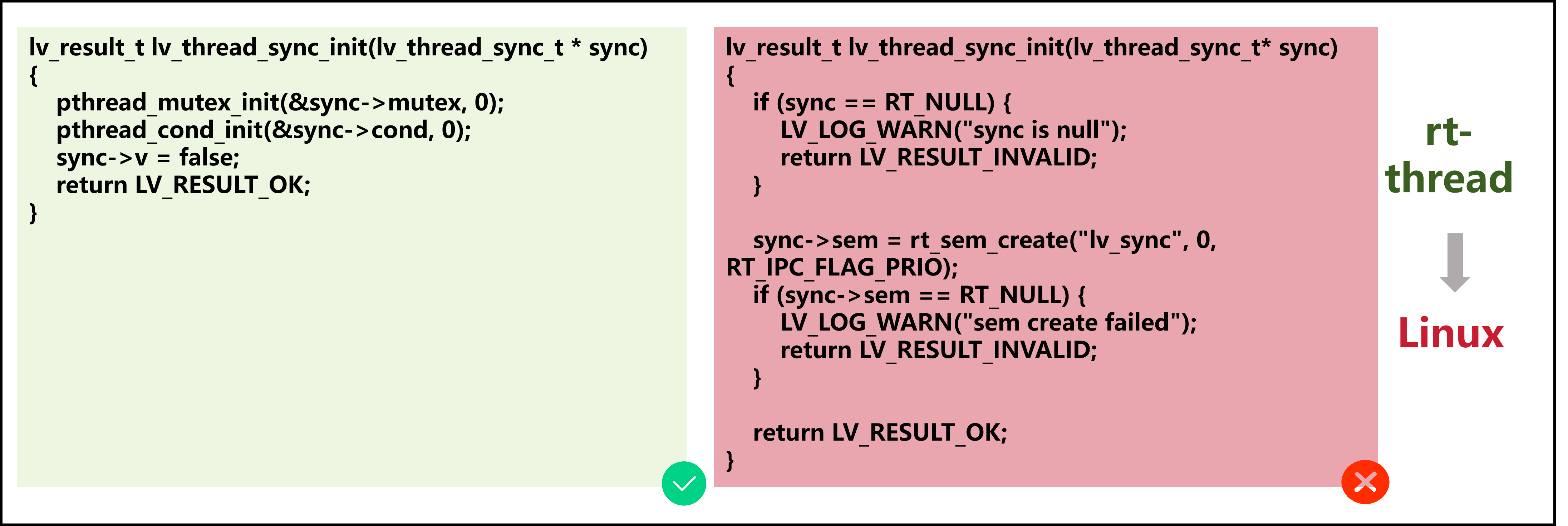}
\caption{Misusing target functions Example}
\label{fig:corss_os_misuse_func}
\end{figure}

\textbf{Misusing target functions.} As shown in Table~\ref{tab:classic_errors}, \textit{misusing target functions} accounts for 11.70\% of the analyzed cases. This error occurs when the model attempts to translate a source-side function call into the target OS, but selects an inappropriate target API that does not faithfully preserve the intended function-level behavior. Unlike omitting available functions, where the correct target API is not used at all, this pattern involves actively translating the source logic into a wrong target-side function. As illustrated in Figure~\ref{fig:corss_os_misuse_func}, the reference implementation on the left initializes the synchronization object using \texttt{pthread\_mutex\_init} and \texttt{pthread\_cond\_init}, which together form a mutex--condition-variable based design. However, the generated code on the right translates this logic into \texttt{rt\_sem\_create}, replacing the original function usage with a semaphore-based API. Although both implementations involve synchronization-related functions, the selected target function does not correspond to the original design and therefore fails to preserve the intended semantics of the source code. This kind of error produces translations that appear reasonable at the API level but are functionally inconsistent with the reference implementation.

\begin{figure}[!t]
\centering
\includegraphics[width=\columnwidth]{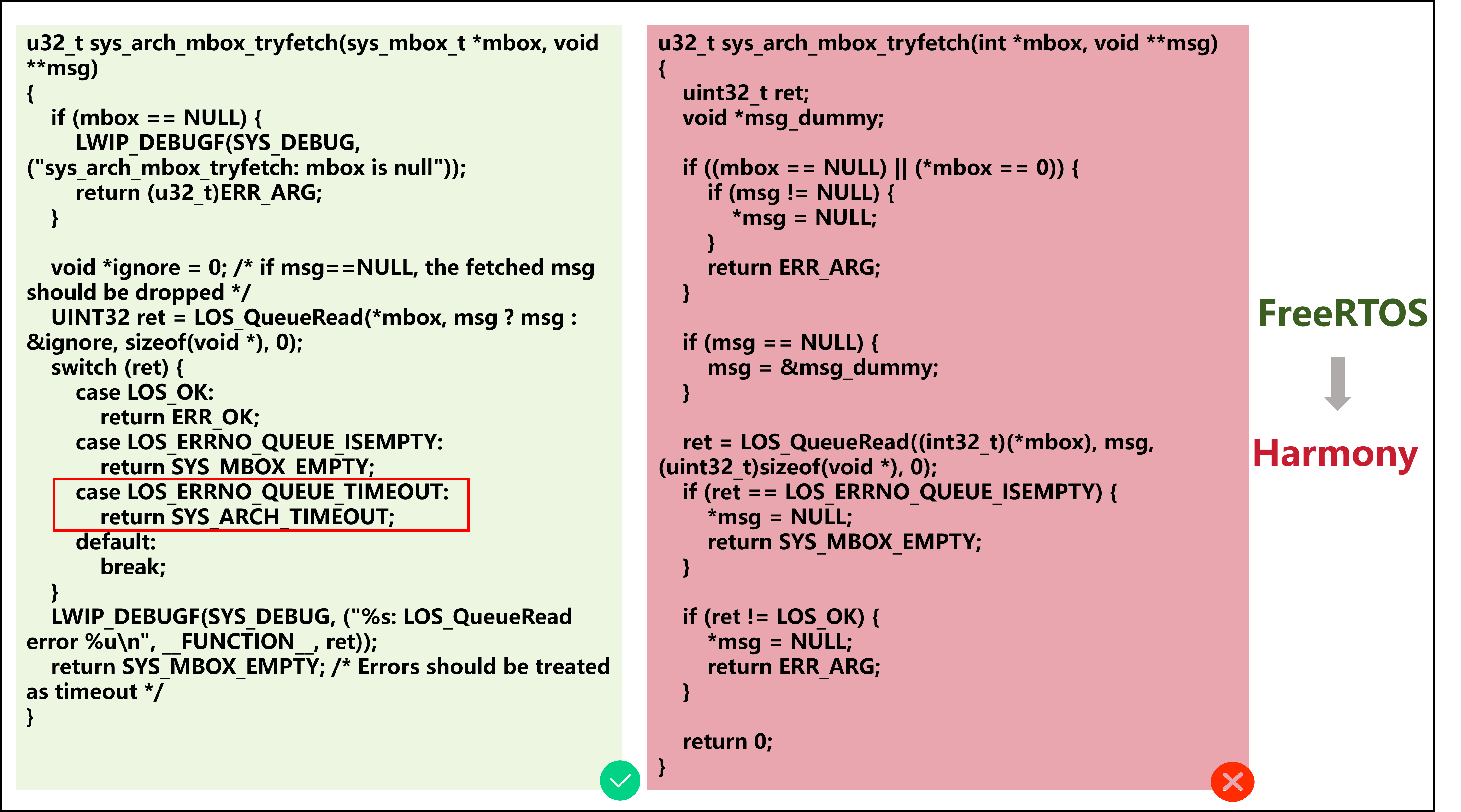}
\caption{Mistranslating Return Values Example}
\label{fig:corss_os_return}
\end{figure}

\textbf{Mistranslating return values.} As shown in Table~\ref{tab:classic_errors}, \textit{mistranslating return values} accounts for 7.80\% of the analyzed cases. This error occurs when the model generates target-side code that does not correctly preserve the return-value semantics of the reference implementation, even if the overall control flow and function usage appear plausible. In cross-OS translation, return values often encode important execution outcomes such as success, empty queue states, timeout conditions, or invalid arguments. Failing to preserve these distinctions can lead to incorrect runtime behavior in the translated code. As illustrated in Figure~\ref{fig:corss_os_return}, in a FreeRTOS-to-Harmony translation example, the reference implementation on the left explicitly distinguishes multiple return cases after calling \texttt{LOS\_QueueRead}, including \texttt{LOS\_OK}, \texttt{LOS\_ERRNO\_QUEUE\_ISEMPTY}, and \texttt{LOS\_ERRNO\_QUEUE\_TIMEOUT}. In particular, the timeout case is correctly mapped to \texttt{SYS\_ARCH\_TIMEOUT}. However, the generated code on the right fails to preserve this return-value branch and does not produce the required \texttt{SYS\_ARCH\_TIMEOUT} result. Instead, it only handles the empty-queue case explicitly and collapses other non-success outcomes into a generic error return. As a result, the translated code loses an important semantic distinction present in the reference implementation and no longer faithfully reflects the intended behavior of the original function.

\section{IMPLICATION}
In this section, we outline three implications inferred from our results to guide the future application and improvement of large language models in embedded software engineering.

\textbf{Implication 1) Embedded software developers should carefully and thoroughly consider the impact of context on LLM performance during code generation, with particular emphasis on the local code window, as it is the most critical for maintaining performance.} Our results show that context substantially affects embedded code generation. Across all six evaluated models, the \texttt{NoCtx} setting causes the largest performance drop, indicating that the target span alone is insufficient for reliable generation. Removing the local code window leads to the largest and most consistent degradation, while removing call dependencies also clearly reduces performance. This suggests that nearby code and execution-related context impose the most direct implementation constraints, helping the model stay consistent with file-level logic, coding patterns, and execution flow. Although global data and type definitions have a smaller average impact, they still provide important semantic support and help reduce deviations from project-specific behavior. Therefore, in practice, embedded developers should treat context construction as a key part of prompt design and prioritize local code, call-chain information, and other project-relevant context whenever possible.

\textbf{Implication 2) When using LLMs for cross-operating-system translation, embedded developers should pay close attention to the source and target OS types as well as common error patterns during translation.}
Our results show that cross-OS translation performance is highly direction-dependent, and even the same OS pair can differ greatly across opposite directions. This means developers should not view cross-system translation as a symmetric API replacement task, but should instead account for semantic differences between the source and target operating systems. Error analysis further shows that many failures follow recurring patterns, including mismatched semantic granularity, incorrect parameter interpretation, and omission of implicit behavioral constraints. In practice, such errors may produce code that appears reasonable at the surface level but behaves incorrectly under the target OS. Therefore, when using LLMs for embedded system translation, developers should carefully consider target-OS characteristics and remain aware of common translation risks, especially those involving timing semantics, parameter conventions, and hidden execution constraints.

\textbf{Implication 3) LLM trainers should place greater emphasis on training data that connects design semantics with implementation details, especially for state-machine-related embedded tasks.}
The state machine results reveal a clear gap between structural understanding and code generation. Although current models can capture states and transitions to some extent, they perform much worse on implementation-oriented tasks such as \texttt{P2C} and \texttt{R2C}. Our error analysis shows that these failures mainly involve API usage mismatch, implementation structure mismatch, transition logic simplification, and misplaced condition evaluation. This suggests that the main limitation lies not in recognizing state-machine structure, but in translating design intent into project-consistent code. Therefore, model trainers should pay more attention to data that reflects how state machines are implemented in real embedded projects, including implementation skeletons, condition-check placement patterns, transition expansion styles, and project-specific API conventions. Strengthening such data during training could help reduce implementation errors and better align generated code with embedded software practice.

\section{CONCLUSION}
In this paper, we present RealEmbedBench, a benchmark specifically designed to evaluate the capabilities of LLMs in real-world application-layer embedded software development. The motivation for constructing RealEmbedBench arises from the observation that existing benchmarks either focus on general-purpose coding tasks or emphasize low-level hardware and driver interactions, leaving application-layer challenges largely unexplored. On the basis of 15 representative open-source projects, we constructed 5,142 manually reviewed samples and systematically evaluated six LLMs. Our results show that project context, especially the local code window, is critical for code generation; state machine tasks reveal persistent gaps between structural understanding and faithful implementation; and cross-OS translation performance strongly depends on the source-target pair. These findings suggest that developers should carefully consider project context and translation direction when using LLMs, while trainers should invest in multi-representation state machine data to improve model reliability and alignment with embedded software practices. We hope RealEmbedBench can facilitate future research on building more reliable LLMs for real embedded development.


\begin{thebibliography}{1}
\bibliographystyle{IEEEtran}

\bibitem{EmbedAgent}
Xu, R., Cao, J., Wu, M., Li, W., Chen, X., Zhang, J., Wang, Y., EmbedAgent: Benchmarking Large Language Models in Embedded System Development, arXiv preprint, abs/2506.11003 (2025)

\bibitem{englhardt:chi-2024}
Englhardt, Z., Li, R., Nissanka, D., Wang, Y., Zhang, C., Gupta, A., Mahajan, D., Iannucci, B., Rowe, A., Banerjee, S., Exploring and characterizing large language models for embedded system development and debugging, Extended Abstracts of the CHI Conference on Human Factors in Computing Systems, CHI EA '24, pp. 1--9 (2024)

\bibitem{garousi2018testing}
Garousi, V., Felderer, M., Karap{\i}{\c{c}}ak, {\c{C}}. M., Y{\i}lmaz, U., Testing embedded software: A survey of the literature, Information and Software Technology, 104, pp. 14--45 (2018)

\bibitem{schirner2007embedded}
Schirner, G., Sachdeva, G., Gerstlauer, A., D{\"o}mer, R., Embedded software development in a system-level design flow, Embedded System Design: Topics, Techniques and Trends, pp. 289--298 (2007)

\bibitem{eclipse-iot-survey-2024}
Eclipse Foundation, IoT and Embedded Survey Report 2024, White Paper, Eclipse Foundation (2024)

\bibitem{yang2021single}
Yang, N., Cuijpers, P., Schiffelers, R., Lukkien, J., Serebrenik, A., Single-state state machines in model-driven software engineering: an exploratory study, Empirical Software Engineering, 26(6), 124 (2021)

\bibitem{yang2020painting}
Yang, N., Cuijpers, P., Schiffelers, R., Lukkien, J., Serebrenik, A., Painting flowers: Reasons for using single-state state machines in model-driven engineering, Proceedings of the 17th International Conference on Mining Software Repositories, pp. 362--373 (2020)

\bibitem{yang2024embedgenius}
Yang, H., Li, M., Han, M., Li, Z., Xu, W., Embedgenius: Towards automated software development for generic embedded iot systems, arXiv preprint, abs/2412.09058 (2024)

\bibitem{wkasowski2005code}
W{a}sowski, A., Code generation and model driven development for constrained embedded software, PhD thesis, IT University of Copenhagen (2005)

\bibitem{abdelsamea2016real}
Abdelsamea, M. H. A., Zorkany, M., Abdelkader, N., Real time operating systems for the internet of things, vision, architecture and research directions, 2016 World Symposium on Computer Applications \& Research (WSCAR), pp. 72--77 (2016)

\bibitem{collie2020m3}
Collie, B., Ginsbach, P., Woodruff, J., Rajan, A., O'Boyle, M. F. P., M3: Semantic api migrations, Proceedings of the 35th IEEE/ACM International Conference on Automated Software Engineering, pp. 90--102 (2020)

\bibitem{antonio2012systematic}
Antonio, E. A., Ferrari, F. C., Fabbri, S. C. P. F., A systematic mapping of architectures for embedded software, 2012 Second Brazilian Conference on Critical Embedded Systems, pp. 18--23 (2012)

\bibitem{xin2024firmporter}
Xin, M., Wen, H., Deng, L., Li, H., Li, Q., Sun, L., FirmPorter: Porting RTOSes at the binary level for firmware re-hosting, International Conference on Information and Communications Security, pp. 310--331 (2024)

\bibitem{humaneval}
Chen, M., Tworek, J., Jun, H., Yuan, Q., de Oliveira Pinto, H. P., Kaplan, J., Edwards, H., Burda, Y., Joseph, N., Brockman, G., et al., Evaluating large language models trained on code, arXiv preprint, abs/2107.03374 (2021)

\bibitem{austin2021program}
Austin, J., Odena, A., Nye, M., Bosma, M., Michalewski, H., Dohan, D., Jiang, E., Cai, C., Terry, M., Le, Q., et al., Program synthesis with large language models, arXiv preprint, abs/2108.07732 (2021)

\bibitem{hendrycks2021measuring}
Hendrycks, D., Basart, S., Kadavath, S., Mazeika, M., Arora, A., Guo, E., Burns, C., Puranik, S., He, H., Song, D., et al., Measuring coding challenge competence with apps, arXiv preprint, abs/2105.09938 (2021)

\bibitem{lai2023ds}
Lai, Y., Li, C., Wang, Y., Zhang, T., Zhong, R., Zettlemoyer, L., Yih, W.-t., Fried, D., Wang, S., Yu, T., DS-1000: A natural and reliable benchmark for data science code generation, International Conference on Machine Learning, pp. 18319--18345 (2023)

\bibitem{lu2021codexglue}
Lu, S., Guo, D., Ren, S., Huang, J., Svyatkovskiy, A., Blanco, A., Clement, C., Drain, D., Jiang, D., Tang, D., et al., Codexglue: A machine learning benchmark dataset for code understanding and generation, arXiv preprint, abs/2102.04664 (2021)

\bibitem{swebench}
Jimenez, C. E., Yang, J., Wettig, A., Yao, S., Pei, K., Press, O., Narasimhan, K., SWE-bench: Can language models resolve real-world github issues?, arXiv preprint, abs/2310.06770 (2024)

\bibitem{tian2024debugbench}
Tian, R., Ye, Y., Qin, Y., Cong, X., Lin, Y., Pan, Y., Wu, Y., Haotian, H., Weichuan, L., Liu, Z., et al., Debugbench: Evaluating debugging capability of large language models, Findings of the Association for Computational Linguistics: ACL 2024, pp. 4173--4198 (2024)

\bibitem{guo2024codeeditorbench}
Guo, J., Li, Z., Liu, X., Ma, K., Zheng, T., Yu, Z., Pan, D., Li, Y., Liu, R., Wang, Y., et al., Codeeditorbench: Evaluating code editing capability of large language models, arXiv preprint, abs/2404.03543 (2024)

\bibitem{zeng2024coderujb}
Zeng, Z., Wang, Y., Xie, R., Ye, W., Zhang, S., Coderujb: An executable and unified java benchmark for practical programming scenarios, Proceedings of the 33rd ACM SIGSOFT International Symposium on Software Testing and Analysis, pp. 124--136 (2024)

\bibitem{bigcodebench}
Zhuo, T. Y., Vu, M. C., Chim, J., Hu, H., Yu, W., Widyasari, R., Yusuf, I. N. B., Zhan, H., He, J., Paul, I., et al., Bigcodebench: Benchmarking code generation with diverse function calls and complex instructions, arXiv preprint arXiv:2406.15877 (2024)

\bibitem{cocobench}
Yin, W., Sun, T., Yu, Y., Fang, J., Su, G., Wang, J., Wang, Z., Wang, W., Chen, R., Dai, Z., et al., CoCo-Bench: A Comprehensive Code Benchmark For Multi-task Large Language Model Evaluation, arXiv preprint arXiv:2504.20673 (2025)

\bibitem{li2024deveval}
Li, J., Li, G., Zhao, Y., Li, Y., Liu, H., Zhu, H., Wang, L., Liu, K., Fang, Z., Wang, L., et al., Deveval: A manually-annotated code generation benchmark aligned with real-world code repositories, Findings of the Association for Computational Linguistics: ACL 2024, pp. 3603--3614 (2024)

\bibitem{gautam2025refactorbench}
Gautam, D., Garg, S., Jang, J., Sundaresan, N., Moghaddam, R. Z., Refactorbench: Evaluating stateful reasoning in language agents through code, arXiv preprint, abs/2503.07832 (2025)

\bibitem{ceedling_build_system}
ThrowTheSwitch.org, Ceedling: Build System for Unit Tests in C, version 1.0.0, https://github.com/ThrowTheSwitch/Ceedling (2024)

\bibitem{unity_test_framework}
ThrowTheSwitch.org, Unity: Unit Testing for C, version 2.6.0, https://github.com/ThrowTheSwitch/Unity (2024)

\bibitem{deepseek2025deepseek}
DeepSeek-AI, Liu, A., Mei, A., et al., DeepSeek-V3.2: Pushing the Frontier of Open Large Language Models, arXiv preprint, arXiv:2512.02556 (2025)

\bibitem{yang2025qwen3}
Yang, A., Li, A., Yang, B., Zhang, B., Hui, B., Zheng, B., Yu, B., Gao, C., Huang, C., Lv, C., et al., Qwen3 technical report, arXiv preprint, arXiv:2505.09388 (2025)

\bibitem{zhipu2025glm47}
Zhipu AI, GLM-4.7: Advancing Code Intelligence and Reasoning Capabilities, Hugging Face, https://huggingface.co/zai-org/GLM-4.7 (2025)

\bibitem{openai2025gpt52}
OpenAI, Introducing GPT-5.2, https://openai.com/zh-Hans-CN/index/introducing-gpt-5-2/ (2025)

\bibitem{google2025gemini31pro}
Google DeepMind, Gemini 3.1 Pro: Best for complex tasks and bringing creative concepts to life, https://deepmind.google/models/gemini/pro/ (2025)

\bibitem{bigcode2024starcoder215b}
BigCode, Lozhkov, A., Li, R., Ben Allal, L., et al., StarCoder2-15B: A 15B parameter code generation model trained on 600+ programming languages, Hugging Face, https://huggingface.co/bigcode/starcoder2-15b (2024)

\bibitem{ren2020codebleu}
Ren, S., Guo, D., Lu, S., Zhou, L., Liu, S., Tang, D., Sundaresan, N., Zhou, M., Blanco, A., Ma, S., Codebleu: a method for automatic evaluation of code synthesis, arXiv preprint, arXiv:2009.10297 (2020)

\bibitem{wokwi}
Wokwi, https://wokwi.com/ (accessed 2026)

\end{thebibliography}
\end{document}